%% file: paper.tex
\pdfoutput=1

\documentclass[10pt]{article}
\usepackage[top=1.25in,bottom=1.25in,left=1.25in,right=1.25in]{geometry}

\usepackage{algorithm}
\usepackage[noend]{algorithmic}

\usepackage{amssymb,amsmath,amsthm,amsfonts}
\usepackage{mathtools}

\usepackage[utf8]{inputenc} 
\usepackage[T1]{fontenc}    
\usepackage{hyperref}       
\usepackage{url}            
\usepackage{booktabs}       

\usepackage{amsfonts}       
\usepackage{nicefrac}       
\usepackage{microtype}      

\usepackage[%
    minnames=4,maxnames=99,maxcitenames=4,
    style=alphabetic,
    doi=false,url=false,
    firstinits=true,hyperref,natbib,backend=bibtex]{biblatex}
\renewbibmacro{in:}{%
  \ifentrytype{article}{}{\printtext{\bibstring{in}\intitlepunct}}}
\bibliography{paper}

\usepackage[capitalize]{cleveref}

\crefname{lemma}{Lemma}{Lemmas}
\crefname{corollary}{Corollary}{Corollaries}
\crefname{theorem}{Theorem}{Theorems}

\usepackage[usenames,dvipsnames]{xcolor}
\input{commenting.tex}

\definecolor{linenocolor}{rgb}{.85,0.85,.85}

\input{macros}

\newcommand{\toolname}{\textsc{BREAD}}
\newcommand{\middlegiven}{\, \middle| \,}

\def\[#1\]{\begin{align}#1\end{align}}

\newcommand{\INPUT}{\item[\textbf{Input:}]}
\newcommand{\NEXTINPUT}{\item[\hphantom{\textbf{Input:}}]}
\newcommand{\OUTPUT}{\item[\textbf{Output:}]}

\newcommand{\jdiv}{{\mathrm{D}_{\rm J}}}

\title{Measuring the reliability of MCMC inference with bidirectional Monte Carlo}
\date{}  

\author{
  Roger B.~Grosse \\ \small Department of Computer Science \\ \small University of Toronto \\
  \and
  Siddharth Ancha \\ \small Department of Computer Science \\ \small University of Toronto \\
  \and
  Daniel M.~Roy \\ \small Department of Statistics \\ \small University of Toronto}

\begin{document}

\maketitle

\begin{abstract}
Markov chain Monte Carlo (MCMC) is one of the main workhorses of probabilistic inference, but it is notoriously hard to measure the quality of approximate posterior samples. This challenge is particularly salient in black box inference methods, which can hide details and obscure inference failures. In this work, we extend the recently introduced bidirectional Monte Carlo \citep{BDMC} technique to evaluate MCMC-based posterior inference algorithms. By running annealed importance sampling (AIS) chains both from prior to posterior and vice versa on simulated data, we upper bound in expectation the symmetrized KL divergence between the true posterior distribution and the distribution of approximate samples. We present Bounding Divergences with REverse Annealing (\toolname), a protocol for validating the relevance of simulated data experiments to real datasets, and integrate it into two probabilistic programming languages: WebPPL \citep{WebPPL} and Stan \citep{Stan}. As an example of how \toolname~can be used to guide the design of inference algorithms, we apply it to study the effectiveness of different model representations in both WebPPL and Stan.
\end{abstract}

\section{Introduction}

Markov chain Monte Carlo (MCMC) is one of the most important classes of probabilistic inference methods and underlies a variety of approaches to automatic inference \citep[e.g.][]{Bugs,Church,WebPPL,Stan}. Despite its widespread use, it is still difficult to rigorously validate the effectiveness of an MCMC inference algorithm. There are various heuristics for diagnosing convergence (see \cref{sec:related-work}), but reliable quantitative measures are hard to find. This creates difficulties both for end users of automatic inference systems and for experienced researchers who develop models and algorithms.

First, consider the perspective of the end user of an MCMC-based automatic inference system. The user would like to know whether the approximate samples are a good representation of the posterior distribution. She may wish to configure various algorithmic parameters (e.g.~the number of steps for which to run the algorithm), or to choose a problem representation for which inference is effective. Currently, all these choices are hard to make in a systematic way, and standard convergence diagnostics often require significant expertise to interpret. 

Now consider the perspective of a scientist aiming to invent a better MCMC algorithm. It is difficult to measure convergence directly, so instead algorithms are typically compared using a proxy measure, such as the joint likelihood of a state or the probability of held-out observations conditioned on the current state. It would be much more useful to have a single scalar quantity which directly measures the quality of the approximate posterior.

In this paper, we extend the recently proposed bidirectional Monte Carlo (BDMC) \citep{BDMC} method to evaluate certain kinds of MCMC-based inference algorithms by bounding the symmetrized KL divergence (Jeffreys divergence) of approximate samples from the true posterior distribution.  Specifically, our method is applicable to algorithms which can be viewed as importance sampling over an extended state space, such as annealed importance sampling (AIS; \citep{AIS}) or sequential Monte Carlo (SMC; \cite{SMC}). 
BDMC was proposed as a method for accurately evaluating log marginal likelihoods on simulated data, so that these log-likelihood values can be used to benchmark marginal likelihood estimators. We show that it can also be used to measure the accuracy of approximate posterior samples obtained from algorithms like AIS or SMC. More precisely, we refine the analysis of \cite{BDMC} to derive an estimator which upper bounds in expectation the Jeffreys divergence between the distribution of approximate samples and the true posterior distribution. We show that this upper bound is quite accurate on some toy distributions for which both the true Jeffreys divergence and the upper bound can be computed exactly. 

While our method is only directly applicable to certain algorithms such as AIS or SMC, these algorithms involve many of the same design choices as traditional MCMC methods, such as the choice of model representation (e.g.~whether to collapse out certain variables), or the choice of MCMC transition operators. Therefore, the ability to evaluate AIS-based inference should also yield insights which inform the design of MCMC inference algorithms more broadly.

One additional hurdle must be overcome to use BDMC to evaluate posterior inference: the method yields rigorous bounds only for simulated data because it requires an exact posterior sample. One would like to be sure that the results on simulated data accurately reflect the accuracy of posterior inference on the real-world data of interest. We present a protocol, which we term Bounding Divergences with REverse Annealing (\toolname), for using BDMC to diagnose inference quality on real-world data. Specifically, we infer hyperparameters on the real data, simulate data from those hyperparameters, measure inference quality on the simulated data, and validate the consistency of the inference algorithm's behavior between the real and simulated data. (This protocol is somewhat similar in spirit to the parametric bootstrap \citep{bootstrap}.)

We integrate \toolname~into the tool chains of two probabilistic programming languages: WebPPL \cite{WebPPL} and Stan \cite{Stan}. 
Both probabilistic programming systems can be used as automatic inference software packages, 
where the user provides a program specifying a joint probabilistic model over observed and unobserved quantities. 
In principle, probabilistic programming has the potential to put the power of sophisticated probabilistic modeling and efficient statistical inference into the hands of non-experts, but realizing this vision is challenging because
it is difficult for a non-expert user to judge the reliability of results
produced by black-box inference.
We believe \toolname~provides a rigorous, general, and automatic procedure for monitoring the quality of posterior inference, so that the user of a probabilistic programming language can have confidence in the accuracy of the results.
Our approach to evaluating probabilistic programming inference is closely related to independent work \citep{Vikash_subjective}
that is also based on the ideas of BDMC. We discuss the relationships between both methods in \cref{sec:related-work}.

In summary, this work includes four main technical contributions. First, we modify both WebPPL and Stan to implement BDMC. Second, we show that BDMC yields an estimator which upper bounds in expectation the Jeffreys divergence of approximate samples from the true posterior. Third, we present a technique for exactly computing both the true Jeffreys divergence and the upper bound on small examples, and show that the upper bound is often a good match in practice. Finally, we propose the \toolname~protocol for using BDMC to evaluate the accuracy of approximate inference on real-world datasets. We validate \toolname~on a variety of probabilistic models in both WebPPL and Stan. As an example of how \toolname~can be used to guide modeling and algorithmic decisions, we use it to analyze the effectiveness of different representations of a matrix factorization model in both WebPPL and Stan.

\section{Background}

\subsection{WebPPL and Stan}

We focus on two particular probabilistic programming packages. First, we consider WebPPL \citep{WebPPL}, a lightweight probabilistic programming language built on Javascript, and intended largely to illustrate some of the important ideas in probabilistic programming. Inference is based on Metropolis--Hastings (M--H) updates to a program's execution trace, i.e.~a record of all stochastic decisions made by the program. WebPPL has a small and clean implementation, and the entire implementation is described in an online tutorial on probabilistic programming \citep{WebPPL}.

Second, we consider Stan \citep{Stan}, a highly engineered automatic inference system which is widely used by statisticians and is intended to scale to large problems. Stan is based on the No U-Turn Sampler (NUTS; \citep{NoUTurn}), a variant of Hamiltonian Monte Carlo (HMC; \citep{HMC}) which chooses trajectory lengths adaptively. HMC can be significantly more efficient than M--H over execution traces because it uses gradient information to simultaneously update multiple parameters of a model, but is less general because it requires a differentiable likelihood. (In particular, this disallows discrete latent variables unless they are marginalized out analytically.)

\subsection{Annealed Importance Sampling}
\label{sec:ais}

Annealed importance sampling (AIS; \citep{AIS}) is a Monte Carlo algorithm commonly used to estimate (ratios of) normalizing constants. 
More carefully, fix a sequence of $\ndist$ distributions $\pmf_1, \ldots, \pmf_\ndist$, with $\pmf_\distIdx(\genState) = \pmfUnnorm_\distIdx(\genState) / \pfn_\distIdx$.  The final distribution in the sequence, $\pmf_\ndist$, is called the \emph{target} distribution; 
the first distribution, $\pmf_1$, is called the \emph{initial} distribution. It is required that one can obtain one or more exact samples from $\pmf_1$.\footnote{Traditionally, this has meant having access to an exact sampler. However, in this work, we sometimes have access to a \emph{sample} from $\pmf_1$, but not a \emph{sampler}.}
Given a sequence of reversible MCMC transition operators 
$\trans_1, \ldots, \trans_\ndist$, where $\trans_\distIdx$ leaves $\pmf_\distIdx$ invariant,
AIS produces a (nonnegative) unbiased estimate of $\pfn_\ndist / \pfn_1$ as follows: 
first, we sample a random initial state $\genState_1$ from $\pmf_1$ and 
set the initial weight $\weight_1 = 1$. 
For every stage $\distIdx \ge 2$ 
we update the weight $\weight$ and sample the state $\genState_{\distIdx}$ according to
\begin{align}
\weight_\distIdx &\gets \weight_{\distIdx-1}  
                \frac{\pmfUnnorm_{\distIdx}(\genState_{\distIdx-1})}
                       {\pmfUnnorm_{\distIdx - 1}(\genState_{\distIdx-1})}
&
\genState_\distIdx &\gets \text{sample from } \trans_\distIdx \left (\genState \middlegiven \genState_{\distIdx-1} \right). \label{eqn:ais-update}
\end{align}
\citet{AIS} justified AIS by showing that it is a simple importance sampler over an extended state space (see \cref{app:ais_derivation} for a derivation in our notation).
From this analysis, it follows that, after \emph{every} stage $\distIdx$, 
the weight $\weight_{\distIdx}$ is an unbiased estimate of the ratio 
$\pfn_{\distIdx} / \pfn_1$.
Two trivial facts are worth highlighting: 
when $\pfn_1$ is known, $\pfn_1 \weight_{\distIdx}$ is an unbiased estimate of $\pfn_\distIdx$,
and when $\pfn_{\distIdx}$ is known, $\weight_{\distIdx} / \pfn_{\distIdx}$ is an unbiased estimate of $1/\pfn_{1}$.
In practice, it is common to repeat the AIS procedure to produce $\nsamp$ independent estimates and combine these by simple averaging to reduce the variance of the overall estimate.  See \cref{alg:background-ais} for the complete procedure.

\begin{algorithm}[t]
\begin{algorithmic}
\INPUT unnormalized densities $\pmfUnnorm_1,\dotsc,\pmfUnnorm_\ndist$ 
\NEXTINPUT reversible MCMC transition operators $\trans_1,\dotsc,\trans_\ndist$, 
                       where $\trans_\distIdx$ leaves $\pmf_\distIdx = \pmfUnnorm_\distIdx / \pfn_\distIdx$ invariant
\NEXTINPUT $\nsamp$ samples from $\pmf_1 = \pmfUnnorm_1 / \pfn_1$
\OUTPUT unbiased estimate of $\pfn_\ndist / \pfn_1$
	\FOR{$\sampleIdx = 1 \textrm{ to } \nsamp$} 
		\STATE $\genState_1 \gets$ sample from $\pmf_1(\genState)$
		\STATE $\weightS{\sampleIdx} \gets 1$ 
		\FOR{$\distIdx = 2 \textrm{ to } \ndist$} 
			\STATE $\weightS{\sampleIdx} \gets \weightS{\sampleIdx} \frac{\pmfUnnorm_{\distIdx}(\genState_{\distIdx-1})}{\pmfUnnorm_{\distIdx - 1}(\genState_{\distIdx-1})}$
			\STATE $\genState_\distIdx \gets$ sample from $\trans_\distIdx\left(\genState \, \middle| \, \genState_{\distIdx-1}\right)$
		\ENDFOR
	\ENDFOR
        \RETURN $\pfnRatioEstimateFwd = \frac 1 \nsamp \sum_{\sampleIdx=1}^\nsamp \weightS{\sampleIdx}$
\end{algorithmic}
\caption{Annealed Importance Sampling (AIS)}
\label{alg:background-ais}
\end{algorithm}

In most applications of AIS, the normalization constant $\pfn_{\ndist}$ for the target distribution $\pmf_{\ndist}$ is the focus of attention, and the initial distribution $\pmf_1$ is chosen to have a known normalization constant $\pfn_1$.
Any sequence of intermediate distributions satisfying a mild domination criterion suffices to produce a valid estimate, but in typical applications,
the intermediate distributions are simply defined to be geometric averages
$\pmfUnnorm_\distIdx(\genState) = \pmfUnnorm_1(\genState)^{1-\invTemp_\distIdx}\pmfUnnorm_\ndist(\genState)^{\invTemp_\distIdx}$, 
where the $\invTemp_\distIdx$ are monotonically increasing parameters with $\invTemp_1 = 0$ and $\invTemp_\ndist = 1$. (An alternative approach is to average moments \citep{MomentAveraging}.)

In the setting of Bayesian posterior inference over parameters $\params$ and latent variables $\latent$ given some fixed observation $\obs$,
we take $\pmfUnnorm_1 (\params,\latent) = \pmf(\params, \latent)$ to be the prior distribution
(hence $\pfn_1 = 1$), 
and we take $\pmfUnnorm_\ndist(\params, \latent) = \pmf(\params, \latent, \obs) = \pmf(\params, \latent)\, \pmf(\obs \given \params, \latent)$. This can be viewed as the unnormalized posterior distribution, whose normalizing constant $\pfn_T = \pmf(\obs)$ is the marginal likelihood.
Using geometric averaging, the intermediate distributions are then
\begin{align}
\pmfUnnorm_\distIdx(\params, \latent) = \pmf(\params, \latent)\, \pmf(\obs \given \params, \latent)^{\invTemp_\distIdx}.
\end{align}
In addition to moment averaging, reasonable intermediate distributions can be produced in the Bayesian inference setting by  conditioning on a sequence of increasing subsets of data; this insight relates AIS to the seemingly different class of sequential Monte Carlo (SMC) methods \citep{SMC}.

\subsection{Stochastic lower bounds on the log partition function ratio}
\label{sec:lower_bounds}

AIS produces a nonnegative unbiased estimate $\pfnRatioEstimateFwd$
of the ratio $\pfnRatioFwd = \pfn_\ndist / \pfn_1$ of partition functions.
Unfortunately, because such ratios often vary across many orders of magnitude, it frequently happens that $\pfnRatioEstimateFwd$ underestimates $\pfnRatioFwd$ with overwhelming probability, while occasionally taking extremely large values. Correspondingly, the variance may be extremely large, or even infinite.

For these reasons,  it is more meaningful to estimate $\log \pfnRatioFwd$.
Unfortunately, the logarithm of a nonnegative unbiased estimate (such as the AIS estimate) is, in general, a biased estimator of the log estimand. 
More carefully, let $\hat A$ be a nonnegative unbiased estimator for $A = \expect[\hat A]$.
Then,
by Jeffreys inequality, 
$
\expect[\log \hat A ] \leq \log \expect[\hat A] = \log A,     
$
and so $\log \hat A $ is a lower bound on $\log A$ in expectation.
The estimator $\log \hat A$ satisfies another important property: 
by Markov's inequality for nonnegative random variables, 
$
\Pr ( \log \hat A > \log A + b ) < e^{-b},   
$
and so $\log \hat A$ is extremely unlikely to overestimate $\log A$ by any appreciable number of nats.
These observations motivate the following definition \citep{BurdaEtAl}:
a \emph{stochastic lower bound} on $X$ is an estimator $\hat X$ satisfying 
$\expect[\hat X] \le X$ and $\Pr ( \hat X > X + b ) < e^{-b}$. Stochastic upper bounds are defined analogously.
The above analysis shows that 
$\log \hat A$ is a stochastic lower bound on $\log A$ when $\hat A$ is a nonnegative unbiased estimate of $A$, and, in particular, $\log \pfnRatioEstimateFwd$ is a stochastic lower bound on $\log \pfnRatioFwd$. (It is possible to strengthen the tail bound by combining multiple samples \cite{Vibhav}.)

\subsection{Reverse AIS and Bidirectional Monte Carlo}
\label{sec:bdmc}

Upper and lower bounds are most useful in combination, as one can then sandwich the true value. As described above, AIS produces a stochastic lower bound on the ratio $\pfnRatioFwd$; many other algorithms do as well. Upper bounds are more challenging to obtain.
The key insight behind bidirectional Monte Carlo (BDMC; \citep{BDMC}) is that,  
\emph{provided one has an exact sample from the target distribution $\pmf_\ndist$},
one can run AIS \emph{in reverse} to produce a stochastic lower bound  
on $\log \pfnRatioRev = \log \pfn_1 / \pfn_\ndist$, 
and therefore a stochastic \emph{upper} bound on $\log \pfnRatioFwd = -\log \pfnRatioRev$. (In fact, BDMC is a more general framework which allows a variety of partition function estimators, but we focus on AIS for pedagogical purposes.)

\begin{algorithm}[t]
\begin{algorithmic}
\INPUT unnormalized densities $\pmfUnnorm_1,\dotsc,\pmfUnnorm_\ndist$ 
\NEXTINPUT reversible MCMC transition operators $\trans_1,\dotsc,\trans_\ndist$, 
                       where $\trans_\distIdx$ leaves $\pmf_\distIdx = \pmfUnnorm_\distIdx / \pfn_\distIdx$ invariant
\NEXTINPUT sample $\genState_1$ from $\pmf_1 = \pmfUnnorm_1 / \pfn_1$ and
                       sample $\genState_\ndist$ from $\pmf_\ndist = \pmfUnnorm_\ndist / \pfn_\ndist$ 
\OUTPUT stochastic lower and upper bounds on $\log \frac {\pfn_\ndist} {\pfn_1}$
   \STATE $\pfnRatioEstimateRev \gets$ AIS 
                        on $\pmfUnnorm_\ndist,\dotsc, \pmfUnnorm_1$; $\trans_T, \dotsc, \trans_1$; and $\genState_\ndist$
   \STATE $\pfnRatioEstimateFwd \gets$ AIS 
                       on $\pmfUnnorm_1,\dotsc,\pmfUnnorm_\ndist$; $\trans_1, \dotsc, \trans_\ndist$; and $\genState_1$
  \RETURN $(\log \pfnRatioEstimateFwd,\, \log \pfnRatioEstimateRev^{-1})$
\end{algorithmic}
\caption{Bidirectional Monte Carlo (BDMC)}
\label{alg:bdmc}
\end{algorithm}

More carefully, for $\distIdx = 1,\dotsc,T$,
define $\tilde \pmf_{\distIdx} = \pmf_{\ndist-\distIdx+1}$
and $\tilde \trans_{\distIdx} = \trans_{\ndist-\distIdx+1}$.
Then $\tilde \pmf_{1}$ corresponds to our original target distribution $\pmf_{\ndist}$ 
and $\tilde \pmf_{\ndist}$ corresponds to our original initial distribution $\pmf_{1}$. 
As before, $\tilde \trans_{\distIdx}$ leaves $\tilde \pmf_{\distIdx}$ invariant.
Consider the estimate produced by AIS on the sequence of distributions $\tilde \pmf_{1},\dotsc,\tilde \pmf_{\ndist}$ and corresponding MCMC transition operators $\tilde \trans_{1},\dotsc,\tilde \trans_{\ndist}$. (In this case, the forward chain of AIS corresponds to the reverse chain described in \cref{sec:ais}.)
The resulting estimate $\pfnRatioEstimateRev$ is a nonnegative unbiased estimator of $\pfnRatioRev$.
It follows that $\log \pfnRatioEstimateRev$ is a stochastic lower bound on $\log \pfnRatioRev$,
and therefore 
$\log \pfnRatioEstimateRev^{-1}$ 
is a stochastic upper bound on 
$\log \pfnRatioFwd = \log \pfnRatioRev^{-1}$. BDMC is simply the combination of this stochastic upper bound with the stochastic lower bound of \cref{sec:lower_bounds}. (See \cref{alg:bdmc} for pseudocode.) Because AIS is a consistent estimator of the partition function ratio under the assumption of ergodicity \cite{AIS}, the two bounds converge as $\ndist \rightarrow \infty$; therefore, given enough computation, BDMC can sandwich $\log \pfnRatioFwd$ to arbitrary precision.

Returning to the setting of Bayesian inference, given some fixed observation $\obs$, 
we can apply BDMC provided we have exact samples from both the prior distribution $ \pmf(\params, \latent)$ and the posterior distribution $\pmf(\params, \latent | \obs)$.  In practice, the prior is typically easy to sample from, but it is typically infeasible to generate exact posterior samples.  However, in models where we can tractably sample from the joint distribution 
$\pmf(\params, \latent, \obs)$,
we can generate exact posterior samples for \emph{simulated} observations using the elementary fact that 
\[
\pmf (\obs)\, \pmf(\params, \latent | \obs) = \pmf(\params, \latent, \obs) = \pmf(\params, \latent)\, \pmf(\obs | \params, \latent).
\]
In other words, if one ancestrally samples $\params$, $\latent$, and $\obs$, this is equivalent to first generating a dataset $\obs$ and then sampling $(\params, \latent)$ exactly from the posterior. Therefore, for simulated data, one has access to a single exact posterior sample; this is enough to obtain stochastic upper bounds on $\log \pfnRatioFwd = \log \pmf(\obs)$.

\section{Methods}

There are at least two criteria we would desire from a sampling-based approximate inference algorithm in order that its samples be representative of the true posterior distribution: we would like the approximate distribution $\pmfProposal(\params, \latent ; \obs)$ to cover all the high-probability regions of the posterior $\pmf(\params, \latent \given \obs)$, and we would like it to avoid placing probability mass in low-probability regions of the posterior. The former criterion motivates measuring the KL divergence $\kldiv( \pmf(\params, \latent \given \obs) \klBars \pmfProposal(\params, \latent ; \obs) )$, and the latter criterion motivates measuring $\kldiv( \pmfProposal(\params, \latent ; \obs) \klBars \pmf(\params, \latent \given \obs) )$. If we desire both simultaneously, this motivates paying attention to the Jeffreys divergence, defined as $\jdiv(q\|p) = \kldiv(q\|p) + \kldiv(p\|q)$.

\begin{figure}
\begin{center}
\begin{small}
(a) \includegraphics[width=0.21 \textwidth]{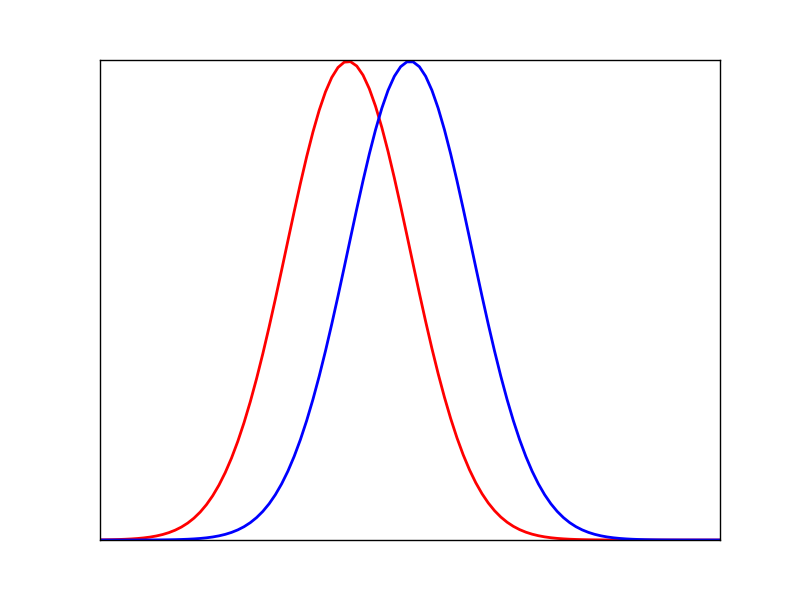}
(b) \includegraphics[width=0.21 \textwidth]{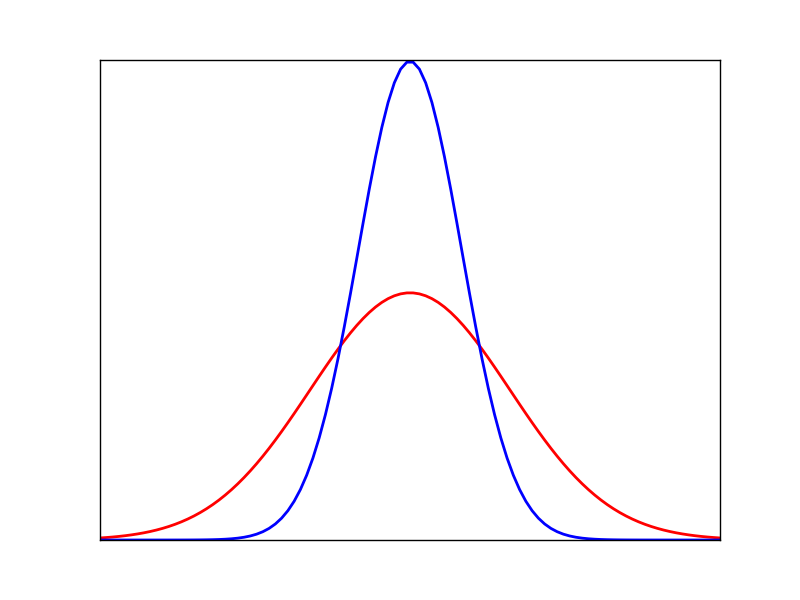}
(c) \includegraphics[width=0.21 \textwidth]{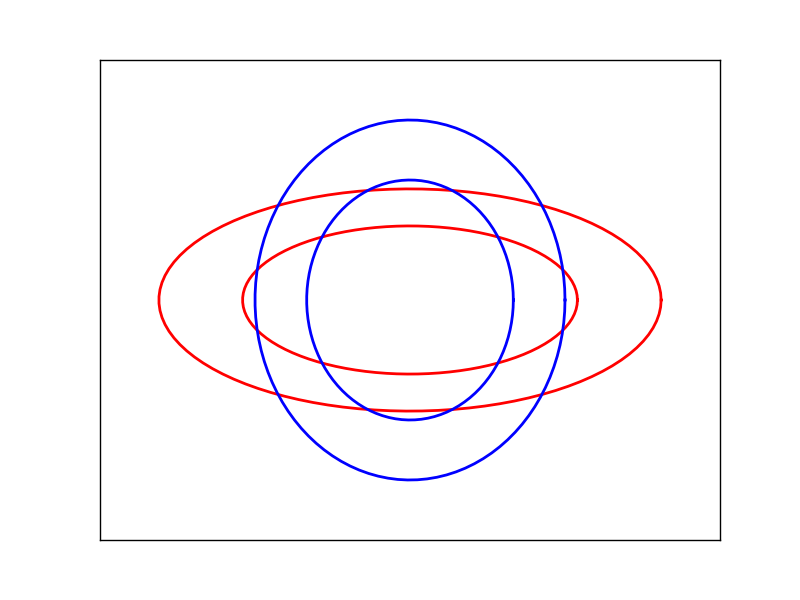}
(d) \includegraphics[width=0.21 \textwidth]{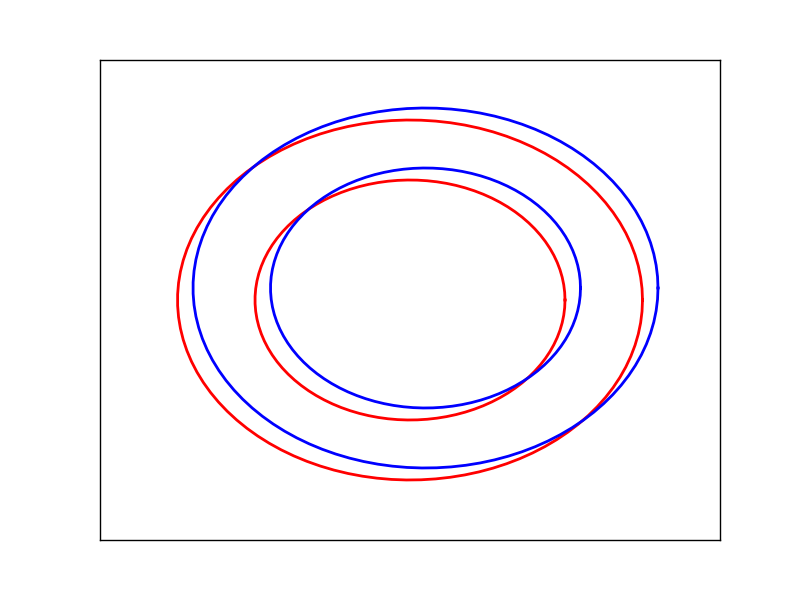}
\end{small}
\end{center}
\caption{Examples of pairs of Gaussian distributions $p$ and $q$ such that $\jdiv(q \klBars p) = 1$. {\bf (a-b)} univariate Gaussians {\bf (c)} bivariate Gaussians {\bf (d)} projection of 100-dimensional Gaussians onto the first two dimensions.}
\label{fig:kl_examples}
\end{figure}

In case one is not accustomed to thinking about the error in approximate posterior inference in terms of the Jeffreys divergence, \cref{fig:kl_examples} provides some examples of pairs of Gaussian distributions whose Jeffreys divergence is 1 nat.

In this section, we present Bounding Divergences with Reverse Annealing (\toolname), a technique for using BDMC to bound the Jeffreys divergence from the true posterior on simulated data, combined with a protocol for using this technique to analyze sampler accuracy on real-world data.

\subsection{Upper bounding the Jeffreys divergence in expectation}
\label{sec:upper_bound_symm}

We now present our technique for bounding the Jeffreys divergence between the target distribution and the distribution of approximate samples produced by AIS. In describing the algorithm, we revert to the abstract state space formalism of \cref{sec:ais}, since the algorithm itself does not depend on any structure specific to posterior inference (except for the ability to obtain an exact sample). We first repeat the derivation from \citep{BDMC} of the bias of the stochastic lower bound $\log \pfnRatioEstimateFwd$. Let $\auxiliary = (\genState_1, \ldots, \genState_{\ndist - 1})$ denote all of the variables sampled in AIS \emph{before the final stage}; the final state $\genState_\ndist$ corresponds to the approximate sample produced by AIS. We can write the distributions over the forward and reverse AIS chains as:
\begin{align}
\pmfProposalForward(\auxiliary, \genState_\ndist) &= \pmfProposalForward(\auxiliary)\, \pmfProposalForward(\genState_\ndist \given \auxiliary) \\
\pmfProposalBackward(\auxiliary, \genState_\ndist) &= \pmf_\ndist(\genState_\ndist)\, \pmfProposalBackward(\auxiliary \given \genState_\ndist).
\end{align}
The distribution of approximate samples $\pmfProposalForward(\genState_\ndist)$ is obtained by marginalizing out $\auxiliary$. Note that sampling from $\pmfProposalBackward$ requires sampling exactly from $\pmf_\ndist$, so strictly speaking, \toolname\ is limited to those cases where one has at least one exact sample from $\pmf_\ndist$ --- such as simulated data from a probabilistic model (see \cref{sec:bdmc}).

The expectation of the estimate $\log \pfnRatioEstimateFwd$ of the log partition function ratio is given by:
\begin{align}
\expect[\log \pfnRatioEstimateFwd] &= \expect_{\pmfProposalForward(\auxiliary, \genState_\ndist)} \left[ \log \frac{\pmfUnnorm_\ndist(\genState_\ndist)\, \pmfProposalBackward(\auxiliary \given \genState_\ndist)}{\pfn_1 \, \pmfProposalForward(\auxiliary, \genState_\ndist)} \right] \\
&= \log \pfn_\ndist - \log \pfn_1 - \kldiv(\pmfProposalForward(\genState_\ndist)\, \pmfProposalForward(\auxiliary \given \genState_\ndist) \klBars \pmf_\ndist(\genState_\ndist)\, \pmfProposalBackward(\auxiliary \given \genState_\ndist)) \label{eqn:KLupper_chain} \\
&\leq \log \pfn_\ndist - \log \pfn_1 - \kldiv(\pmfProposalForward(\genState_\ndist) \klBars \pmf_\ndist(\genState_\ndist)). \label{eqn:KLupper}
\end{align}
(Note that $\pmfProposalForward(\auxiliary \given \genState_\ndist)$ is the \emph{conditional} distribution of the forward chain, given that the final state is $\genState_\ndist$.) The inequality follows because marginalizing out variables cannot increase the KL divergence.

We now go beyond the analysis in \citep{BDMC}, to bound the bias in the other direction. The expectation of the reverse estimate $\pfnRatioEstimateRev$ is
\begin{align}
\expect[\log \pfnRatioEstimateRev] &= \expect_{\pmfProposalBackward(\genState_\ndist, \auxiliary)} \left[ \log \frac {\pfn_1 \, \pmfProposalForward(\auxiliary, \genState_\ndist)}{\pmfUnnorm_\ndist(\genState_\ndist) \, \pmfProposalBackward(\auxiliary \given \genState_\ndist)} \right] \\
&= \log \pfn_1 - \log \pfn_\ndist - \kldiv( \pmf_\ndist(\genState_\ndist) \, \pmfProposalBackward (\auxiliary | \genState_\ndist)
                \klBars 
                \pmfProposalForward(\genState_\ndist) \,  \pmfProposalForward(\auxiliary \given \genState_\ndist)
              ) \label{eqn:revKLupper_chain} \\
&\leq \log \pfn_1 - \log \pfn_\ndist - \kldiv( \pmf_\ndist(\genState_\ndist) 
                \klBars 
                \pmfProposalForward(\genState_\ndist)
              ). \label{eqn:revKLupper}
\end{align}

As discussed above, $\log \pfnRatioEstimateFwd$ and $\log \pfnRatioEstimateRev^{-1}$ can both be seen as estimators of $\log \frac{\pfn_\ndist}{\pfn_1}$, the former of which is a stochastic lower bound, and the latter of which is a stochastic upper bound. Consider the gap between these two bounds, $\symmKLUpperBoundEstimate \triangleq \log \pfnRatioEstimateRev^{-1} - \log \pfnRatioEstimateFwd$. It follows from \cref{eqn:KLupper,eqn:revKLupper} that, in expectation, $\symmKLUpperBoundEstimate$ upper bounds the Jeffreys divergence
\begin{equation}  
\symmKLTrue \triangleq \jdiv(\pmf_\ndist(\genState_\ndist) , \pmfProposalForward(\genState_\ndist))
\triangleq 
    \kldiv ( \pmf_\ndist(\genState_\ndist) 
                  \klBars 
                \pmfProposalForward(\genState_\ndist) )
     + 
\kldiv (     \pmfProposalForward(\genState_\ndist)
                  \klBars 
           \pmf_\ndist(\genState_\ndist)  )
\end{equation}
between the target distribution $\pmf_\ndist$ and the distribution $\pmfProposalForward(\pmf_\ndist)$ of approximate samples.

Alternatively, if one happens to have some other lower bound $\genLowerBound$ or upper bound $\genUpperBound$ on $\log \pfnRatioFwd$, then one can bound either of the one-sided KL divergences by running only one direction of AIS. Specifically, from \cref{eqn:KLupper}, $\expect[\genUpperBound - \log \pfnRatioEstimateFwd] \geq \kldiv(\pmfProposalForward(\genState_\ndist) \klBars \pmf_\ndist(\genState_\ndist))$, and from \cref{eqn:revKLupper}, $\expect[\log \pfnRatioEstimateRev^{-1} - \genLowerBound] \geq \kldiv( \pmf_\ndist(\genState_\ndist) \klBars \pmfProposalForward(\genState_\ndist))$.

\subsection{Evaluating $\symmKLUpperBound$ and $\symmKLTrue$ on small examples}
\label{sec:exact_kl}

In the previous section, we derived an estimator $\symmKLUpperBoundEstimate$ which upper bounds in expectation the Jeffreys divergence $\symmKLTrue$ between the true posterior distribution and the distribution of approximate samples produced by AIS, i.e., 
$\expect[\symmKLUpperBoundEstimate] > \symmKLTrue$. It can be seen from \cref{eqn:KLupper_chain,eqn:revKLupper_chain} that the expectation $\symmKLUpperBound \triangleq \expect[\symmKLUpperBoundEstimate]$ is the Jeffreys divergence between the distributions over the forward and reverse chains:
\begin{align}
\symmKLUpperBound &= \expect[\log \pfnRatioEstimateRev^{-1} - \log \pfnRatioEstimateFwd ] \nonumber \\
&= \kldiv(\pmfProposalForward(\genState_\ndist)\, \pmfProposalForward(\auxiliary \given \genState_\ndist) \klBars \pmf_\ndist(\genState_\ndist)\, \pmfProposalBackward(\auxiliary \given \genState_\ndist)) \ + \nonumber\\
&\phantom{=} + \ \kldiv( \pmf_\ndist(\genState_\ndist) \, \pmfProposalBackward (\auxiliary | \genState_\ndist)
                \klBars 
                \pmfProposalForward(\genState_\ndist) \,  \pmfProposalForward(\auxiliary \given \genState_\ndist)
              ) \nonumber \\
&= \kldiv ( \pmfProposalForward(\auxiliary, \genState_\ndist)
                  \klBars 
                \pmfProposalBackward(\auxiliary, \genState_\ndist) 
              ) + \kldiv ( \pmfProposalBackward(\auxiliary, \genState_\ndist) 
                  \klBars 
                \pmfProposalForward(\auxiliary, \genState_\ndist)
              ) \nonumber \\
 &= \jdiv ( \pmfProposalForward(\auxiliary, \genState_\ndist)
                  \klBars 
                \pmfProposalBackward(\auxiliary, \genState_\ndist)  
              )
\end{align}
One might intuitively expect $\symmKLUpperBound$ to be a very loose upper bound on $\symmKLTrue$, as it is a divergence over a much larger domain (specifically, of size $|\genStateSpace|^\ndist$, compared with $|\genStateSpace|$). In order to test this empirically, we have evaluated both $\symmKLTrue$ and $\symmKLUpperBound$ on several toy distributions for which both quantities can be tractably computed. We found the na{\"\i}ve intuition to be misleading --- on some of these distributions, $\symmKLUpperBound$ is a good proxy for $\symmKLTrue$. In this section, we describe how to exactly compute $\symmKLUpperBound$ and $\symmKLTrue$ in small discrete spaces; experimental results are given in \cref{sec:exact_kl_results}

Even when the domain $\genStateSpace$ is a small discrete set, distributions over the extended state space are too large to represent explicitly. However, because both the forward and reverse chains are Markov chains, all of the marginal distributions $\pmfProposalForward(\genState_{\distIdx}, \genState_{\distIdx+1})$ and $\pmfProposalBackward(\genState_{\distIdx}, \genState_{\distIdx+1})$ can be computed using the forward pass of the Forward--Backward algorithm. The final-state divergence $\symmKLTrue$ can be computed directly from the marginals over $\genState_\ndist$. The KL divergence $\kldiv(\pmfProposalForward \klBars \pmfProposalBackward)$ can be computed as:
\begin{align}
\kldiv(\pmfProposalForward \klBars \pmfProposalBackward) &= \expect_{\pmfProposalForward(\genState_1, \ldots, \genState_\ndist)} \left[ \log \pmfProposalForward(\genState_1, \ldots, \genState_\ndist) - \log \pmfProposalBackward(\genState_1, \ldots, \genState_\ndist) \right] \\
&= \expect_{\pmfProposalForward(\genState_1)} \left[ \log \pmfProposalForward(\genState_1) - \log \pmfProposalBackward(\genState_1) \right] \ + \nonumber \\
&\phantom{=} + \ \sum_{\distIdx=1}^{\ndist-1} \expect_{\pmfProposalForward(\genState_{\distIdx}, \genState_{\distIdx+1})} \left[ \log \pmfProposalForward(\genState_{\distIdx+1} \given \genState_{\distIdx}) - \log \pmfProposalBackward(\genState_{\distIdx+1} \given \genState_{\distIdx}) \right],
\end{align}
and analogously for $\kldiv(\pmfProposalBackward \klBars \pmfProposalForward)$. $\symmKLUpperBound$ is the sum of these quantities.

\subsection{Application to real-world data}
\label{sec:fixed-hyper}

So far, we have focused on the setting of simulated data, where it is possible to obtain an exact posterior sample, and then to rigorously bound the Jeffreys divergence using BDMC. However, we are more likely to be interested in evaluating the performance of inference on real-world data. Heuristically, we can generate simulated data to use as a proxy for the real-world data, but this requires ensuring that the datasets are similar enough that the findings will transfer. We propose to simulate data using model hyperparameters obtained from the real-world dataset of interest, run BDMC on the simulated data, and then validate that forward AIS chains behave similarly between the two datasets.

Often, we informally distinguish between parameters of a statistical model (e.g.~regression weights) and hyperparameters (e.g.~noise variance). So far in this paper, we have not distinguished parameters and hyperparameters, since they are treated the same by Stan, WebPPL, and all of the associated inference algorithms. However, we believe parameters and hyperparameters must be treated differently in the synthetic data generation process and in the reverse AIS chain. In this section we use $\params$ to refer to parameters and $\hyperParams$ to refer to hyperparameters. In Bayesian analysis, hyperparameters are often assigned non-informative or weakly informative priors, in order to avoid biasing the inference. This poses a challenge for \toolname, as datasets generated from hyperparameters sampled from such priors (which are often very broad) can be very dissimilar to real datasets, and hence conclusions from the simulated data may not generalize. 

In order to generate synthetic datasets which better match the real-world dataset of interest, we adopt the following heuristic scheme: we first perform approximate posterior inference on the real-world dataset. Let $\hyperParamsReal$ denote the estimated hyperparameters. We then simulate parameters and data from the forward model $\pmf(\params \given \hyperParamsReal) \pmf(\data \given \params, \hyperParamsReal)$. The forward AIS chain is run on $\data$ in the usual way. However, to obtain the posterior sample for the reverse chain, we first start with $(\hyperParamsReal, \params)$, and then run some number of MCMC transitions which preserve $\pmf(\params, \hyperParams \given \data)$. In general, this will not yield an exact posterior sample, since $\hyperParamsReal$ was not sampled from $\pmf(\hyperParams \given \data)$. However, heuristically $\hyperParamsReal$ ought to be fairly representative of the posterior distribution unless the prior $\pmf(\hyperParams)$ concentrates most of its mass away from $\hyperParamsReal$. If this is the case, then a relatively small number of MCMC steps ought to come close to $\pmf(\params, \hyperParams \given \data)$, and hence the sample can be used as a proxy for the exact posterior sample. We validate this hypothesis in \cref{sec:fixed-hyper-experiments}.

We also recommend validating that the inference procedure behaves similarly on the real-world and simulated data. We aren't aware of a rigorous method for testing this, so we recommend a heuristic approach where one runs the forward AIS chains for varying numbers of steps, computes the log-ML lower bounds, and inspects whether the overall shape of the curves is similar. This ought to highlight cases where one of the datasets is obviously much harder than the other. (We would prefer to have a stronger form of validation than this; this is a topic for future work.)

\section{Related work}
\label{sec:related-work}

Much work has already been devoted to the diagnosis of Markov chain convergence. (See e.g.~\citep{GeyerTutorial}.) In general, a set of MCMC samples $\genStateS{1}, \ldots, \genStateS{\nsamp}$ is used to estimate an expectation $\expect[\statGen(\genState)] = \frac{1}{\nsamp} \sum_{\sampleIdx=1}^\nsamp \statGen(\genStateS{\sampleIdx})$. We would like to achieve an estimate with both low bias and low variance. One often attempts to reduce the bias by discarding all samples until a particular ``burn-in'' time, by which the Markov chain is believed to have mixed sufficiently. Unfortunately, choosing a good burn-in time is difficult because it is hard to diagnose mixing. One commonly used heuristic (also recommended by the Stan manual \citep{StanManual}) is to initialize multiple chains at diverse regions of the state space and to track various statistics of all of these chains over time. One wants to check that the between-chain variance of these statistics is small relative to the within-chain variance. This is not a perfect solution for two reasons: (1) all of the chains may be initialized to the same local mode, so the method would only diagnose mixing within this mode, and (2) the set of statistics may not be sufficiently expressive to detect failures to mix. 

While burn-in is meant to address the problem of bias, other diagnostics are meant to deal with the problem of variance. Consecutive states in a Markov chain are highly correlated with one another, so one cannot obtain low-variance estimates of the statistics unless one runs the chain long enough to obtain near-independent samples. This can be formalized through the notion of effective sample size (ESS), defined as the number of truly independent samples which would result in the same variance which was obtained from the MCMC samples. One can estimate the ESS in practice by estimating autocorrelations between the statistics at various time offsets in a Markov chain. Unfortunately, such diagonstics are heuristic, and can also return misleading results in cases where the chain fails to explore important modes.

A lot of work has also been devoted to automatically configuring parameters of MCMC algorithms. Since it is hard to reliably summarize the performance of an MCMC algorithm, such automatic configuration methods typically rely on method-specific analyses. For instance, \citet{PointTwoThreeFour} famously showed that the optimal acceptance rate of M--H with an isotropic proposal distribution is 0.234 under fairly general conditions. M--H algorithms are sometimes tuned to achieve this acceptance rate, even in situations where the theoretical analysis doesn't hold. An easily measurable performance criterion could likely enable more direct optimization of algorithmic hyperparameters.

Finally, \citet{MackeySteinsMethod} present a method for directly estimating the quality of a set of approximate samples, independently of how those samples were obtained. 
This method has strong guarantees under a strong convexity assumption, but unfortunately this assumption rules out exactly those cases where mixing is most difficult: multimodal or badly conditioned distributions. By contrast, our Jeffreys divergence bound holds regardless of the choice of posterior distribution.

We have recently learned of independent work \citep{Vikash_subjective} which also builds off BDMC to evaluate the accuracy of posterior inference in a probabilistic programming language. In particular, \citet{Vikash_subjective} define an unbiased estimator for a quantity called the \emph{subjective divergence}. The estimator is \emph{equivalent} to BDMC except that the reverse chain is initialized from an arbitrary \emph{reference} distribution, rather than the true posterior. 
In \citep{Vikash_subjective},
the subjective divergence is shown to upper bound the Jeffreys divergence when the true posterior is used; this is equivalent to our analysis in \cref{sec:upper_bound_symm}. 

Much less is known about subjective divergence when the reference distribution is not taken to be the true posterior: 
\citet{Vikash_subjective} give conditions under which the subjective divergence bounds the Jeffreys divergence, but the question of when these conditions hold in practice is not explored. Verifying the conditions in a particular case may require computing or bounding KL divergences which are themselves intractable.
To the best of our knowledge, in the absence of further assumptions, subjective divergence has no provable relationship to KL divergence, and so the measure is best regarded as a heuristic. 
(Indeed, the subjective divergence is not truly a divergence, as it can take negative values.\footnote{This can happen when the reference distribution is especially inaccurate. For instance, consider a case where $x$ and $y$ are constant, and $z \in \{0, 1\}$. Define the model distribution $p(z) = p(z, x)$, inference distribution $q(z) = q(z;x)$, and reference distribution $r(z) = r(z;x)$ such that $p(1) = 0.9$, $q(1) = 0.5$, and $r(1) = 0.1$. Let both $\hat{q}_{\rm IS}$ and $\hat{q}_{\rm HM}$ return the true $q(z;x)$ deterministically. In this case, $\subjectiveDivergence(q(z;x^*) \| p(z|x^*)) \approx -0.879$. })
We believe it is an essential feature of \toolname\ that it 
always bounds  the Jeffreys divergence in expectation
on simulated data and then validates the consistency of the algorithm's behavior between simulated and real data.

\section{Empirical evaluation of \toolname}

We have two motivations for our experiments. Our first set of experiments was intended to validate that \toolname\ can be used to evaluate the accuracy of posterior inference in realistic settings. Next, we used \toolname\ to explore the tradeoffs between two different representations of a matrix decomposition model in both WebPPL and Stan.

\subsection{Validation}

As described above, \toolname\ returns rigorous bounds on the Jeffreys divergence only when the data are sampled from the model distribution. Mathematically, it could potentially give misleading results in three ways. First, the upper bound $\symmKLUpperBound$ may overestimate the true Jeffreys divergence $\symmKLTrue$. Second, results on simulated data may not correspond to results on real-world data if the simulated data is not representative of the real-world data. Finally, the fixed hyperparameter procedure of \cref{sec:fixed-hyper} may not yield a sample sufficiently representative of the true posterior $\pmf(\params, \hyperParams \given \data)$. In this section, we attempted to validate that $\symmKLUpperBound$ is a good proxy for $\symmKLTrue$, that the behavior of the method on simulated data is consistent with that on real data, and that the fixed-hyperparameter samples can be used as a proxy for samples from the posterior.

\subsubsection{How tight is the bound?}
\label{sec:exact_kl_results}

To validate the upper bound $\symmKLUpperBound$ as a measure of the accuracy of an approximate inference engine, we must first check that it accurately reflects the true Jeffreys divergence $\symmKLTrue$. This is hard to do in realistic settings since $\symmKLTrue$ is generally unavailable, but we can evaluate both quantities on small toy distributions using the technique of \cref{sec:exact_kl}.

\begin{figure}
\begin{center}
\begin{tabular}{rccccc}
& $\beta = 0.1$ & $\beta = 0.25$ & $\beta = 0.5$ & $\pmf(\genStateUni)$ & $\pmfProposal(\genStateUni)$ \\
\raisebox{0.275in}{\small EASY RANDOM} & 
  \includegraphics[width=0.13 \textwidth]{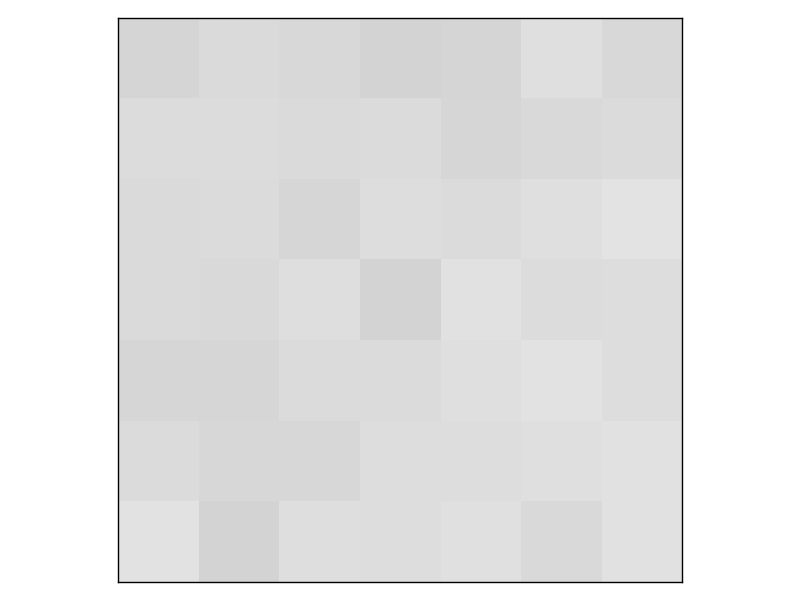} &
  \includegraphics[width=0.13 \textwidth]{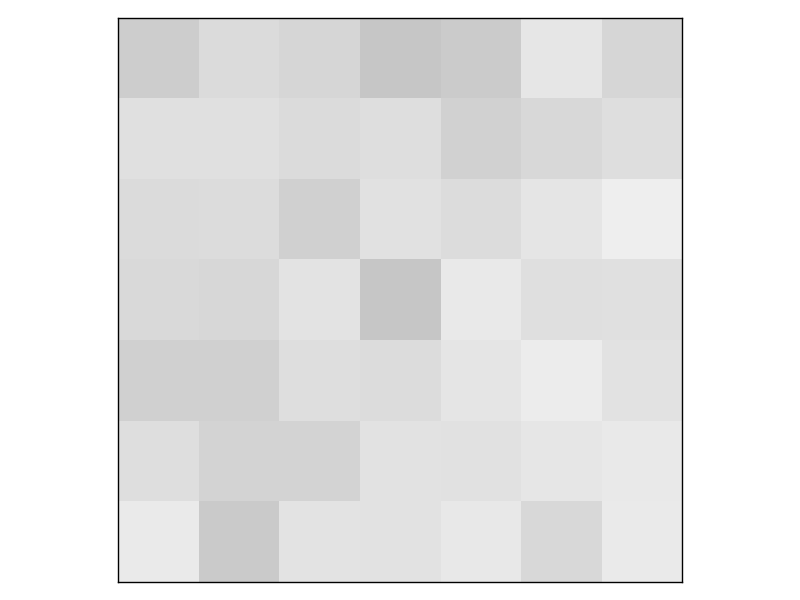} &
  \includegraphics[width=0.13 \textwidth]{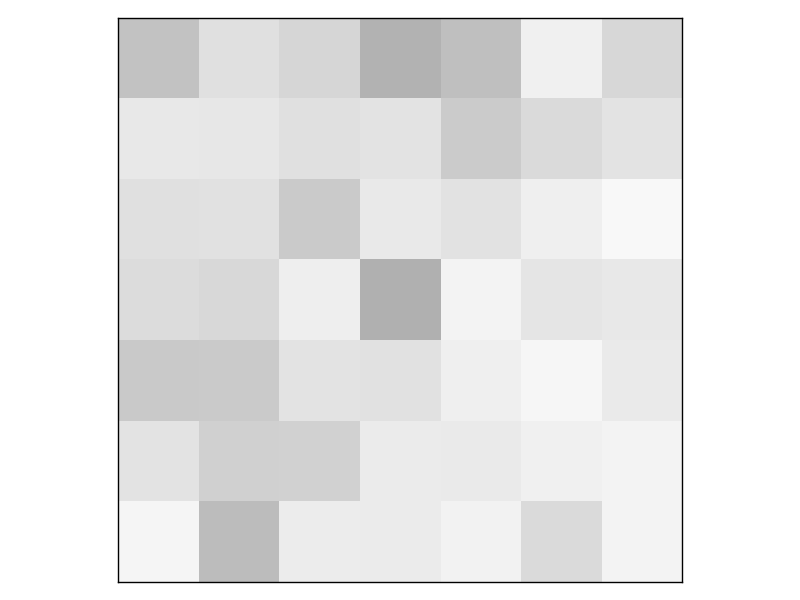} &
  \includegraphics[width=0.13 \textwidth]{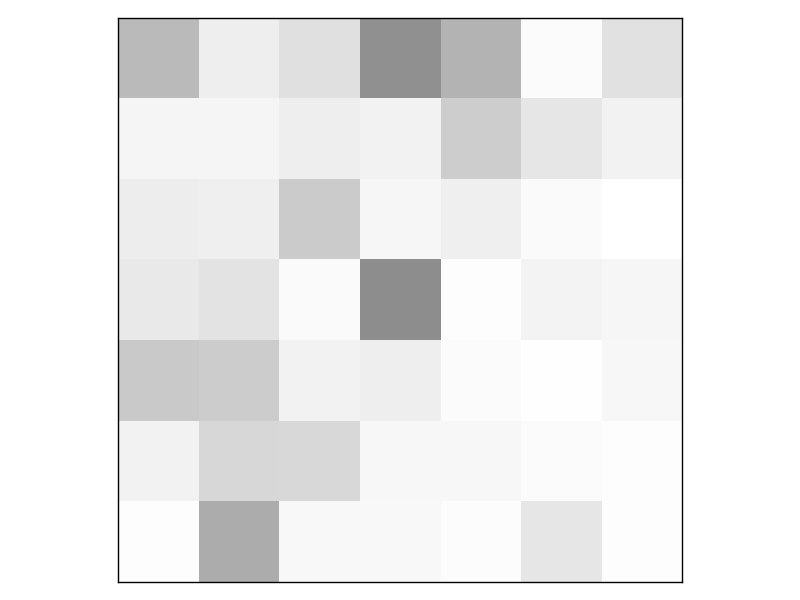} &
  \includegraphics[width=0.13 \textwidth]{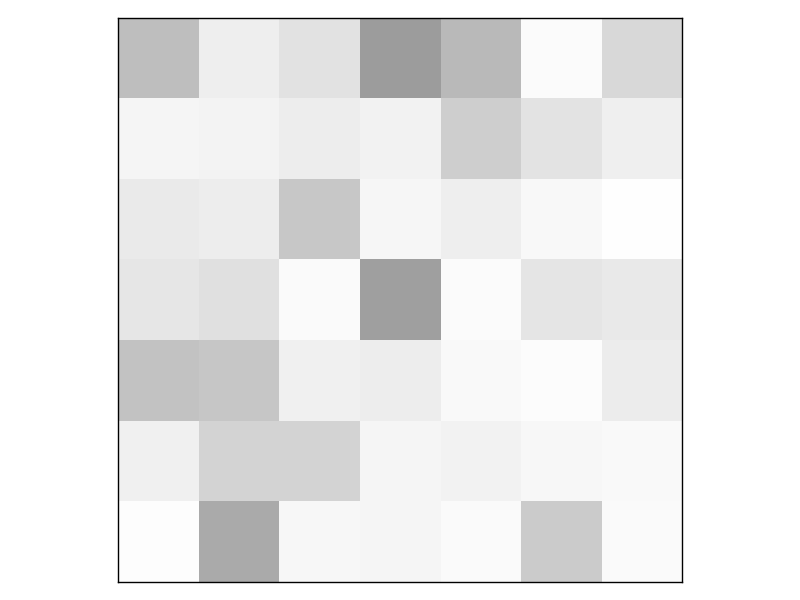} \\
\raisebox{0.275in}{\small HARD RANDOM} &
  \includegraphics[width=0.13 \textwidth]{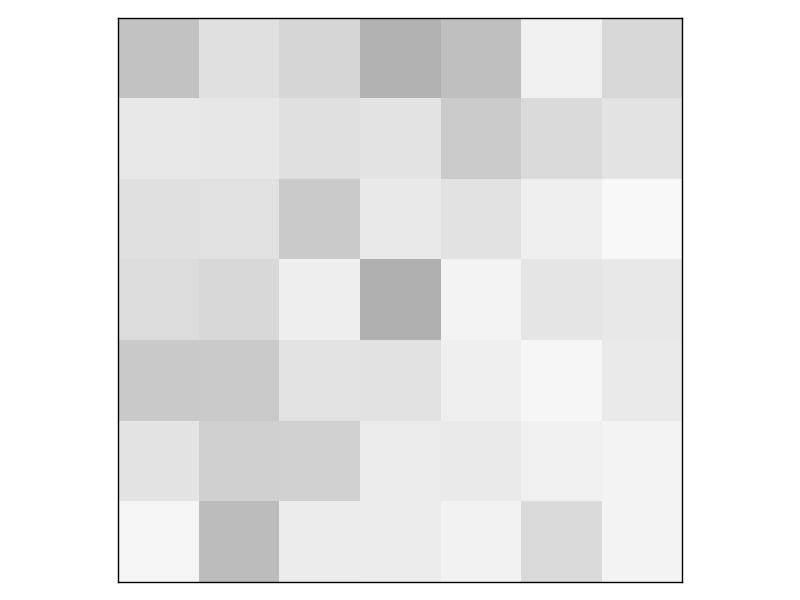} &
  \includegraphics[width=0.13 \textwidth]{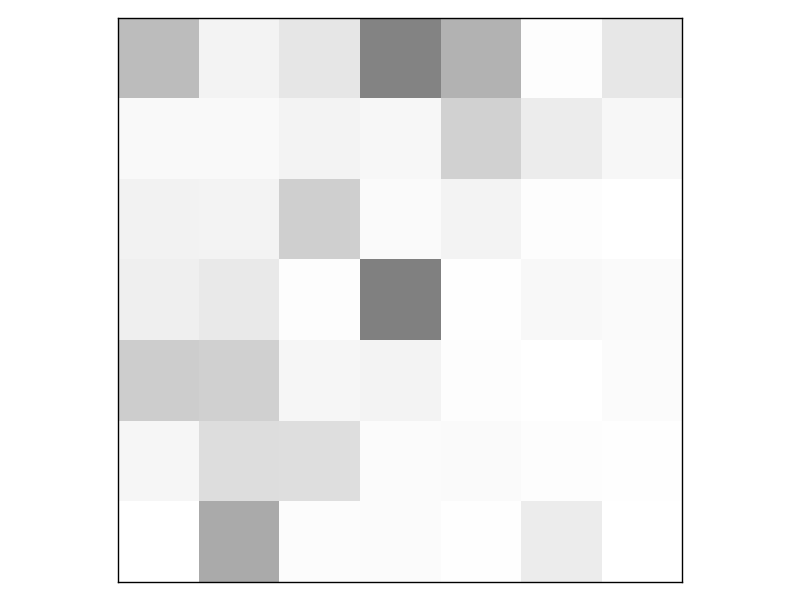} &
  \includegraphics[width=0.13 \textwidth]{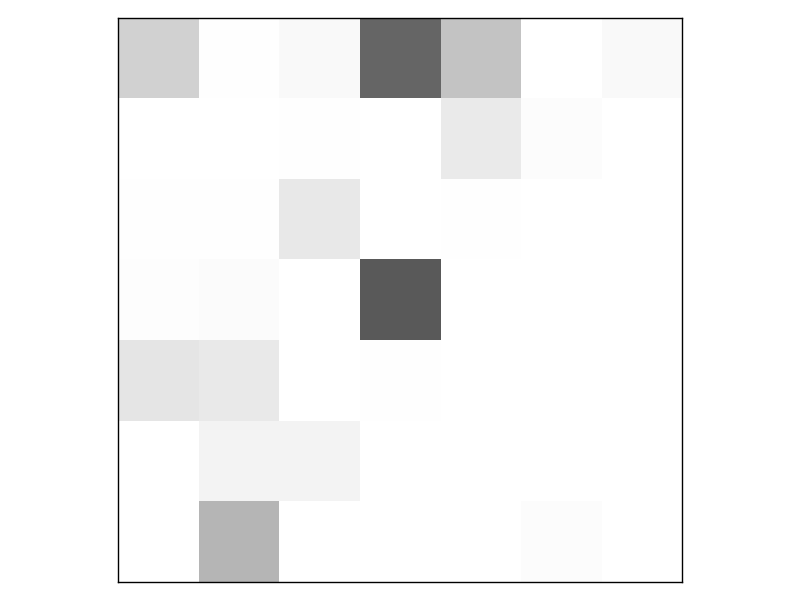} &
  \includegraphics[width=0.13 \textwidth]{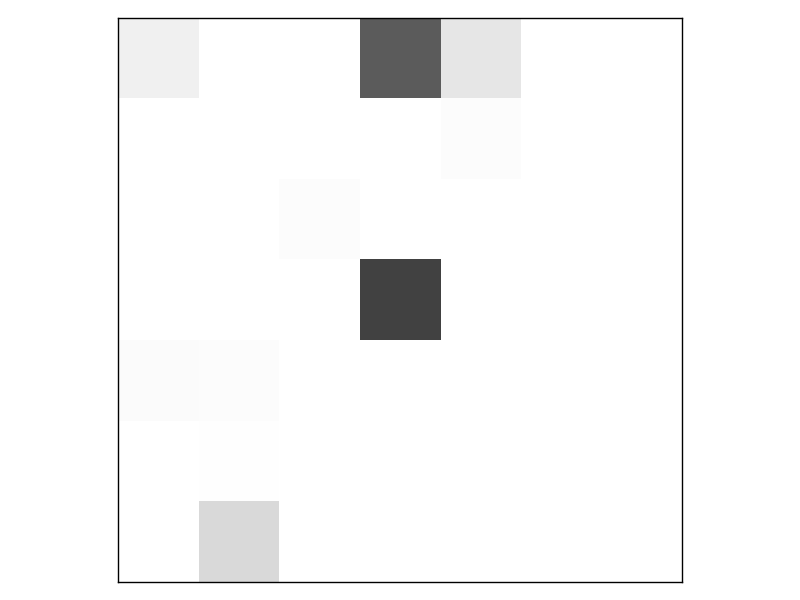} &
  \includegraphics[width=0.13 \textwidth]{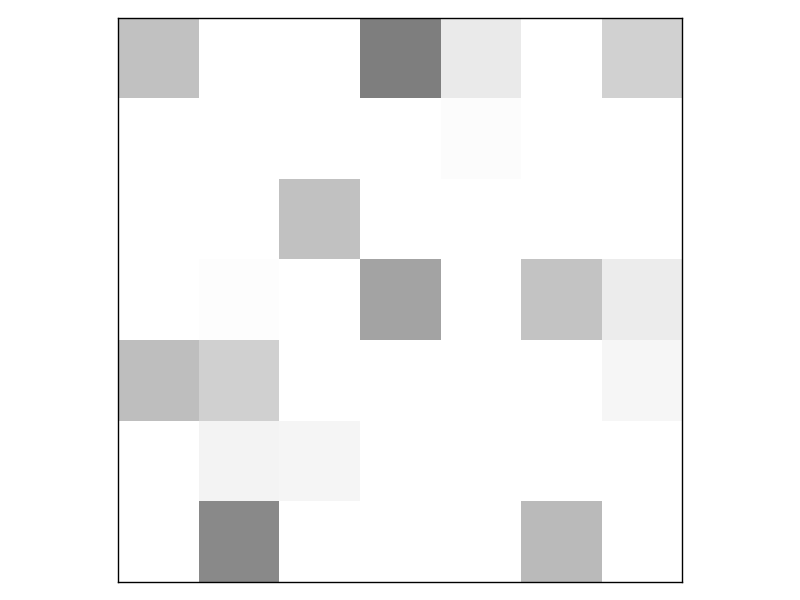} \\
\raisebox{0.275in}{\small BARRIER} &
  \includegraphics[width=0.13 \textwidth]{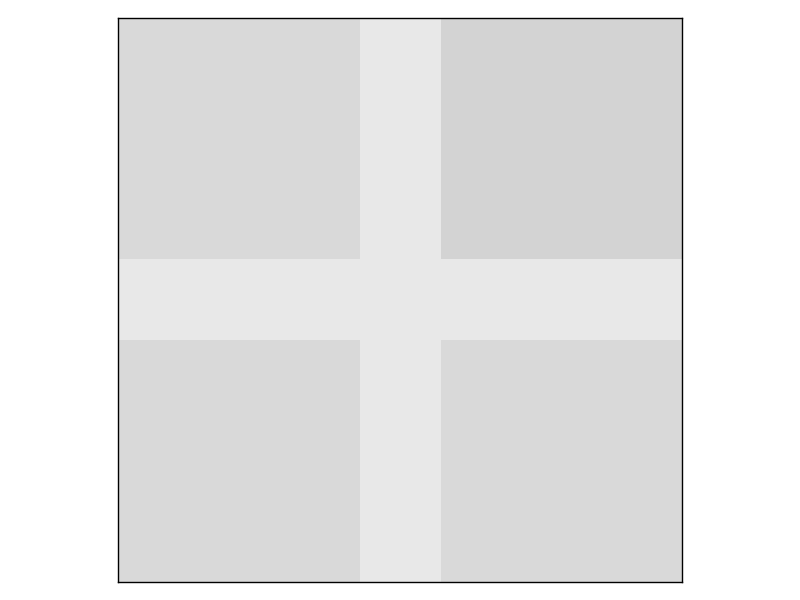} &
  \includegraphics[width=0.13 \textwidth]{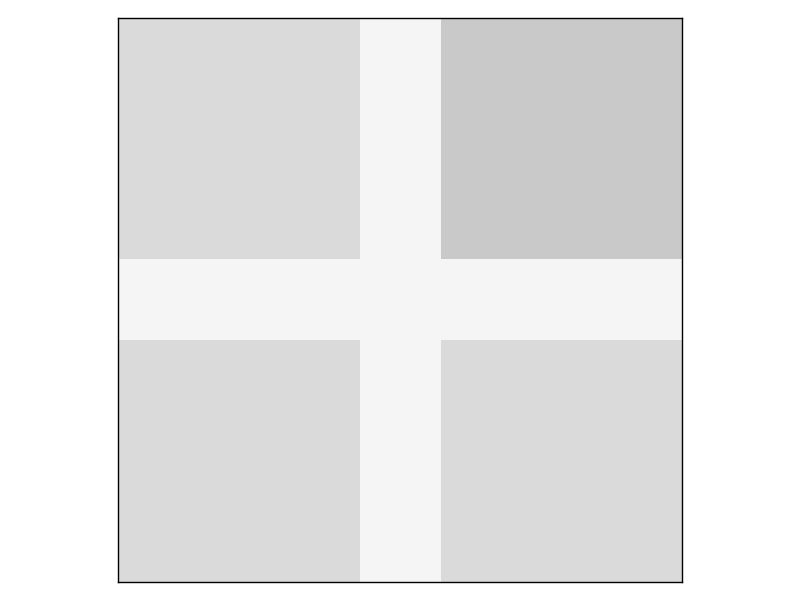} &
  \includegraphics[width=0.13 \textwidth]{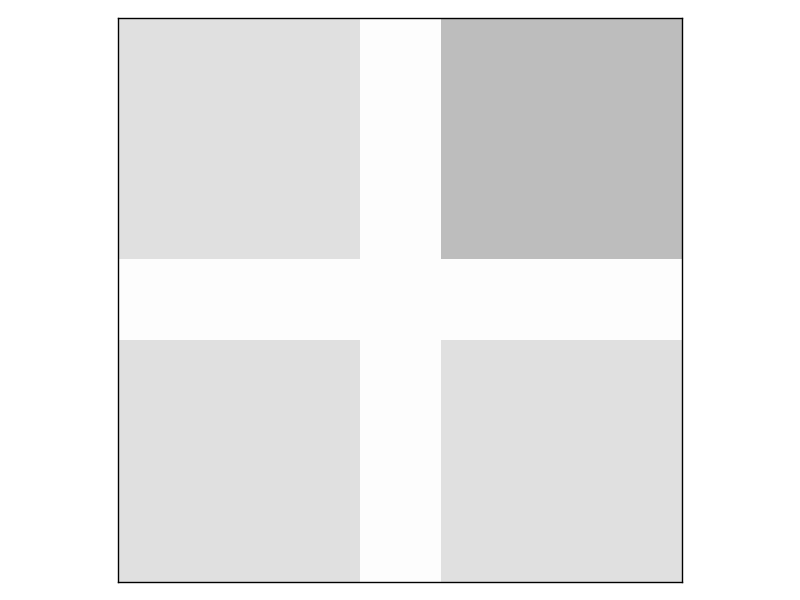} &
  \includegraphics[width=0.13 \textwidth]{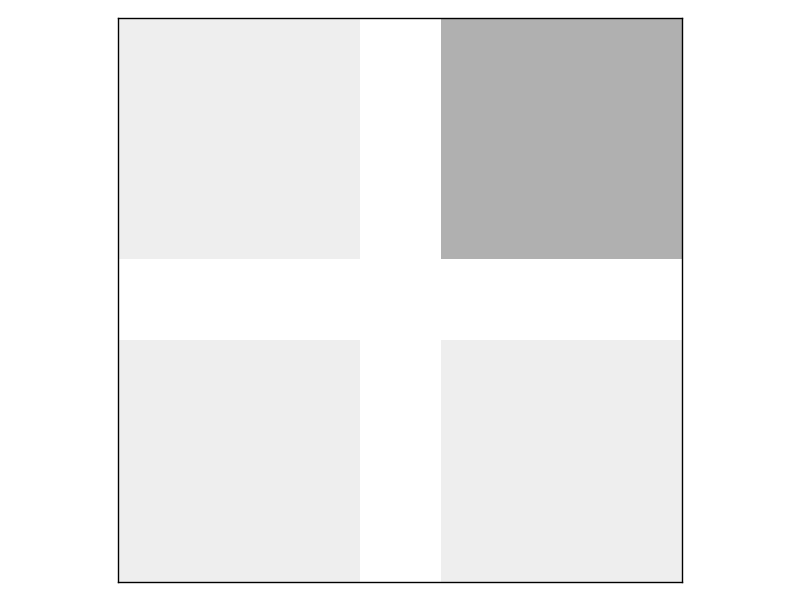} &
  \includegraphics[width=0.13 \textwidth]{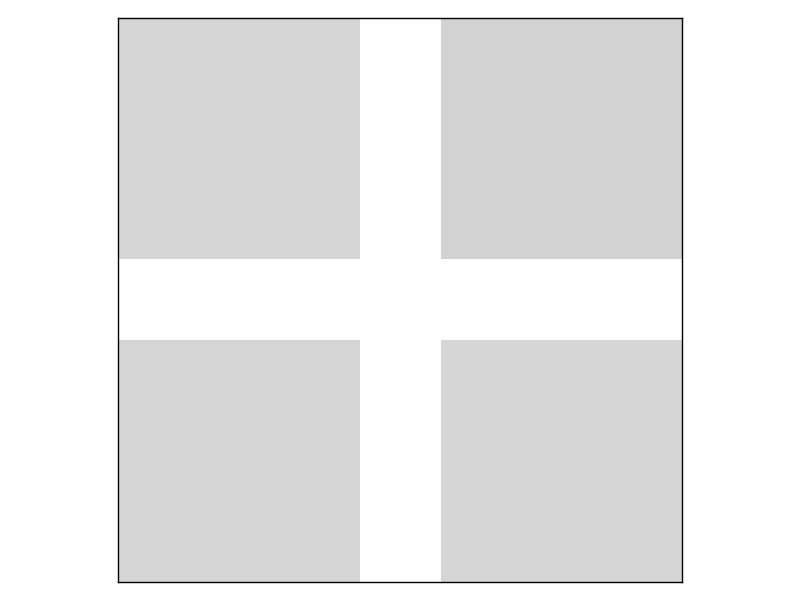} \\
\end{tabular}
\end{center}
\caption{Visualizations of the toy distributions used for analyzing the tightness of our bound. {\bf Columns 1-3:} intermediate distributions for various inverse temperatures $\invTemp$. {\bf Column 4:} target distribution. {\bf Column 5:} Distribution of approximate samples produced by AIS with 100 intermediate distributions.}
\label{fig:toy_dist}
\end{figure}

We considered several toy distributions, where the domain $\genStateSpace$ was taken to be a $7 \times 7$ grid. In all cases, the MCMC transition operator was Metropolis-Hastings, where the proposal distribution was uniform over moving one step in the four coordinate directions. (Moves which landed outside the grid were rejected.) For the intermediate distributions, we used geometric averages with a linear schedule for $\invTemp$.

First, we generated random unnormalized target distributions $\pmfUnnorm(\genStateUni) = \exp(g(\genStateUni))$, where each entry of $g$ was sampled independently from the normal distribution $\normal(0, \sigma)$. When $\sigma$ is small, the M-H sampler is able to explore the space quickly, because most states have similar probabilities, and therefore most M-H proposals are accepted. However, when $\sigma$ is large, the distribution fragments into separated modes, and it is slow to move from one mode to another using a local M-H kernel. We sampled random target distributions using $\sigma=2$ (which we refer to as EASY RANDOM) and $\sigma=10$ (which we refer to as HARD RANDOM). The target distributions, as well as some of the intermediate distributions, are shown in \cref{fig:toy_dist}.

The other toy distribution we considered consists of four modes separated by a deep energy barrier. The unnormalized target distribution is defined by:
\begin{equation}
\pmfUnnorm(\genStateUni) = \left\{ \begin{array}{ll}
    e^3    & \textrm{in the upper right quadrant} \\
    1      & \textrm{in the other three quadrants} \\
    e^{-10} & \textrm{on the barrier}
\end{array} \right.                                   
\end{equation}
In the target distribution, the dominant mode makes up about $87\%$ of the probability mass. \cref{fig:toy_dist} shows the target function as well as some of the AIS intermediate distributions. This example illustrates a common phenomenon in AIS, whereby the distribution splits into different modes at some point in the annealing process, and then the relative probability mass of those modes changes. Because of the energy barrier, we refer to this example as BARRIER.

The distributions of approximate samples produced by AIS with $\ndist=100$ intermediate distributions are shown in \cref{fig:toy_dist}. For EASY RANDOM, the approximate distribution is quite accurate, with $\symmKLTrue = 0.13$. However, for the other two examples, the probability mass is misallocated between different modes because the sampler has a hard time moving between them. Correspondingly, the Jeffreys divergences are $\symmKLTrue = 4.73$ and $\symmKLTrue = 1.65$ for HARD RANDOM and BARRIER, respectively. 

\begin{figure}
\begin{center}
\includegraphics[width=0.5 \textwidth]{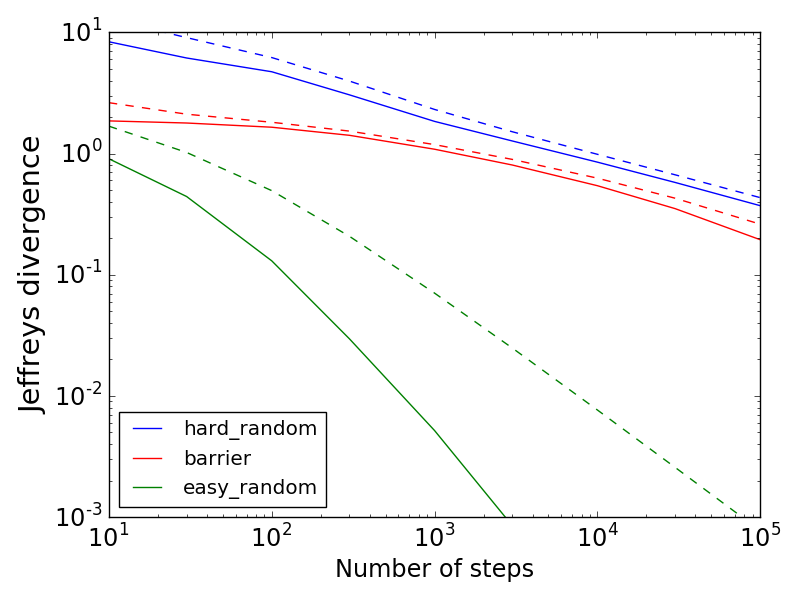}
\end{center}
\caption{Comparing the true Jeffreys divergence $\symmKLTrue$ against our upper bound $\symmKLUpperBound$ for the toy distributions of \cref{fig:toy_dist}. {\bf Solid lines:} $\symmKLTrue$. {\bf Dashed lines:} $\symmKLUpperBound$.}
\label{fig:toy_kl_plot}
\end{figure}

How well are these quantities estimated by the upper bound $\symmKLUpperBound$?  \cref{fig:toy_kl_plot} shows both $\symmKLTrue$ and $\symmKLUpperBound$ for all three toy distributions and numbers of intermediate distributions varying from 10 to 100,000. In the case of EASY RANDOM, the bound is quite far off: for instance, with 1000 intermediate distributions, $\symmKLTrue \approx 0.00518$ and $\symmKLUpperBound \approx 0.0705$, which differ by more than a factor of 10. However, the bound is more accurate in the other two cases: for HARD RANDOM, $\symmKLTrue \approx 1.840$ and $\symmKLUpperBound \approx 2.309$, a relative error of 26\%; for BARRIER, $\symmKLTrue \approx 1.085$ and $\symmKLUpperBound \approx 1.184$, a relative error of about 9\%. In these cases, according to \cref{fig:toy_kl_plot}, the upper bound remains accurate across several orders of magnitude in the number of intermediate distributions and in the true divergence $\symmKLTrue$. Roughly speaking, it appears that most of the difference between the forward and reverse AIS chains is explained by the difference in distributions over the final state. Even the extremely conservative bound in the case of EASY RANDOM is potentially quite useful, as it still indicates that the inference algorithm is performing well, in contrast with the other two cases.

Overall, we conclude that on these toy distributions, the upper bound $\symmKLUpperBound$ is accurate enough to give meaningful information about $\symmKLTrue$. This motivates using $\symmKLUpperBound$ as a diagnostic for evaluating approximate inference algorithms when $\symmKLTrue$ is unavailable.

\subsubsection{Validation on real datasets}
\label{sec:valid-real}

We consider a Bayesian linear regression model in Stan, with weights being sampled from uniform Gaussian priors and the standard deviation of the generating Gaussian distribution sampled from an inverse-Gaussian prior. We consider two datasets - a) a real dataset - a standardized version of the NHEFS dataset (\url{http://cdn1.sph.harvard.edu/wp-content/uploads/sites/1268/2015/07/nhefs_book.xlsx}) and b) a simulated dataset. In order to validate the use of the synthetic data as a proxy for the real data, we generated the stochastic lower bound curves on both datasets. Results are shown in \cref{fixed-param}(a). The curves appear to level off to different values, as different datasets would have different log marginal likelihoods. However, they appear to level off at about the same rate, suggesting that the behavior is consistent, i.e.~neither of the datasets challenges the sampler more than the other. This suggests that bounds on the KL divergence, estimated from the upper and lower bounds on the simulated data, can help us understand convergence properties on the real data as well.

\subsubsection{Fixed hyperparameter scheme}
\label{sec:fixed-hyper-experiments}

\begin{figure}
\begin{center}
\begin{small}
(a) \includegraphics[width=0.33 \textwidth]{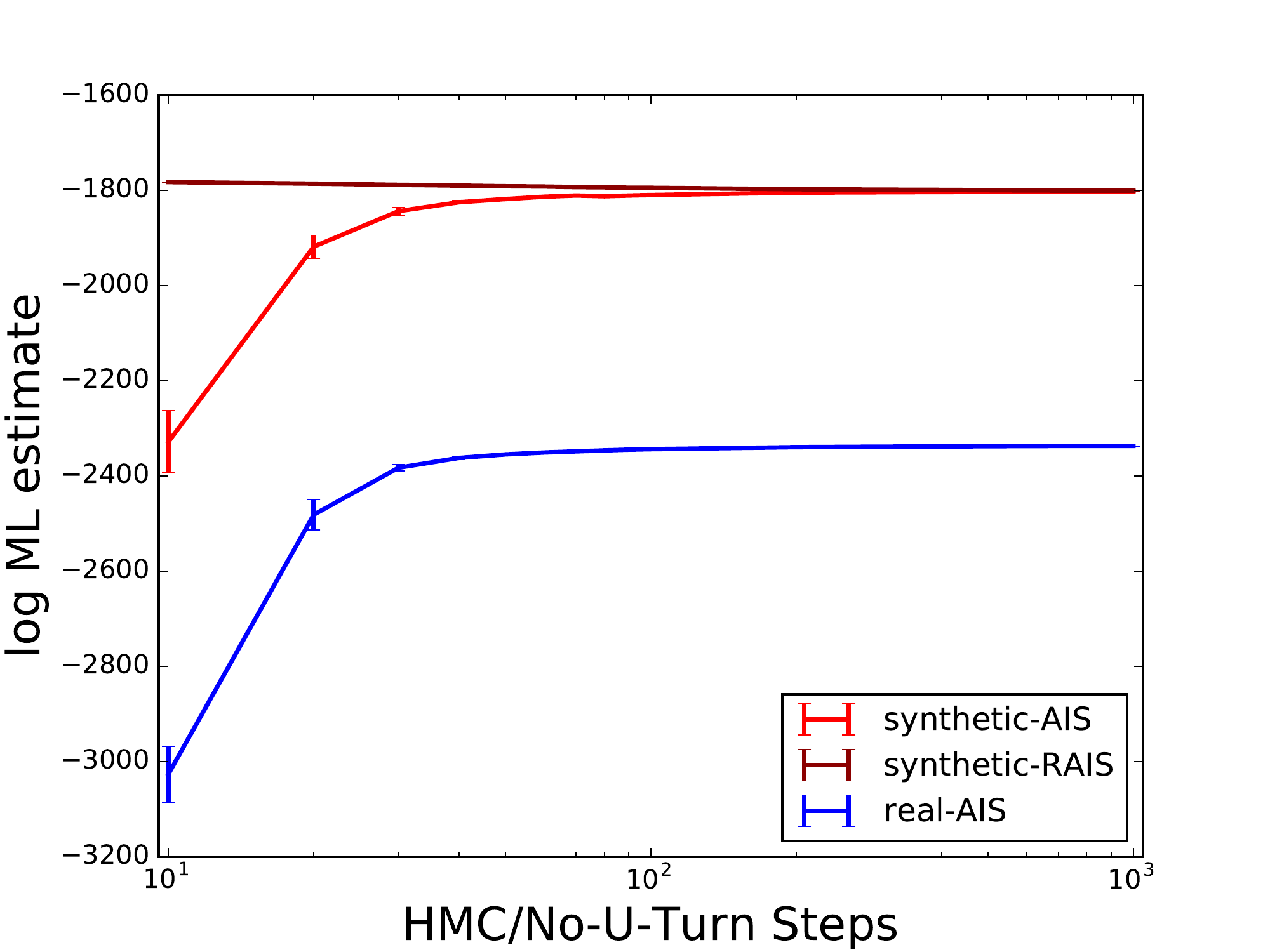}
(b) \includegraphics[width=0.33 \textwidth]{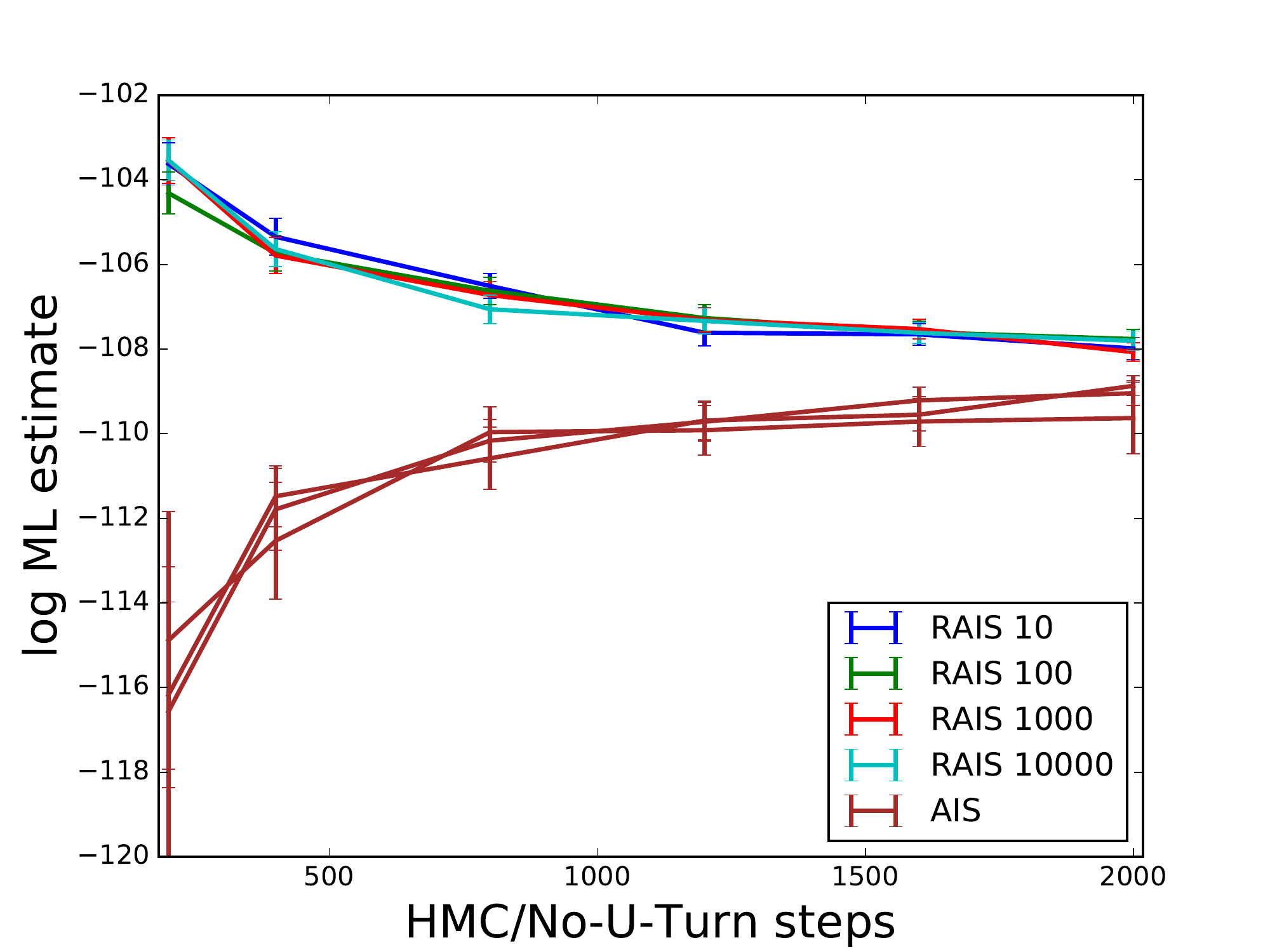}
\end{small}
\end{center}
\caption{{\bf (a)} Forward and reverse AIS on the simulated data, and forward AIS on real-world data, for the Bayesian regression model. The behavior of AIS is consistent between the simulated and real-world data. 
{\bf (b)} Comparison of the stochastic upper bounds obtained from reverse AIS, starting with various numbers of MCMC steps from the true hyperparameters. There is no noticeable difference between these conditions, suggesting that even a small number of MCMC steps gives a reasonable proxy to the true posterior. For comparison, we also show independent estimates of the stochastic lower bound from the forward chain. }
\label{fixed-param}
\end{figure}

To validate the fixed hyperparameter scheme of \cref{sec:fixed-hyper}, we consider a similar Bayesian regression model as in \cref{sec:valid-real}, with covariance hyperparameters assigned broad Cauchy priors. We fixed the hyperparameter values to 1 when simulating data. As described in \cref{sec:fixed-hyper}, we ran MCMC chains of length 10, 100, 1000 and 10,000 starting from $(\hyperParamsReal, \params)$ in order to obtain an approximate posterior sample to initialize the reverse chain. Results are shown in \cref{fixed-param}(b). We find that the upper bound curves are indistinguishable for all numbers of steps. In fact, the variability was observed to be comparable to between-trial variability of independent and identical AIS runs, as seen in \cref{fixed-param}(b). This suggests that reverse AIS chains can be initialized from true hyperparameter values (instead of the posterior), followed by a small number of MCMC steps. Hence \toolname\ supports non-informative or weakly informative priors on hyperparameters.

\subsection{Scientific findings produced by \toolname}

Having validated various aspects of \toolname, we applied it to investigate the choice of model representation in Stan and WebPPL. In the course of our experiments, we also uncovered a bug in WebPPL, indicating the potential usefulness of \toolname\ as a means of testing the correctness of an implementation.

\subsubsection{Comparing model representations}
Many models can be written in more than one way, for example by introducing or collapsing latent variables. Performance of probabilistic programming languages can be sensitive to such choices of representation, and the representation which gives the best performance may vary from one language to another. Hence, users of probabilistic programming languages might want to know the right representation to use for their language of choice, and \toolname\ can prove extremely useful in this regard.

\begin{figure}
\begin{center}
\begin{small}
(a) \includegraphics[width=0.28 \textwidth]{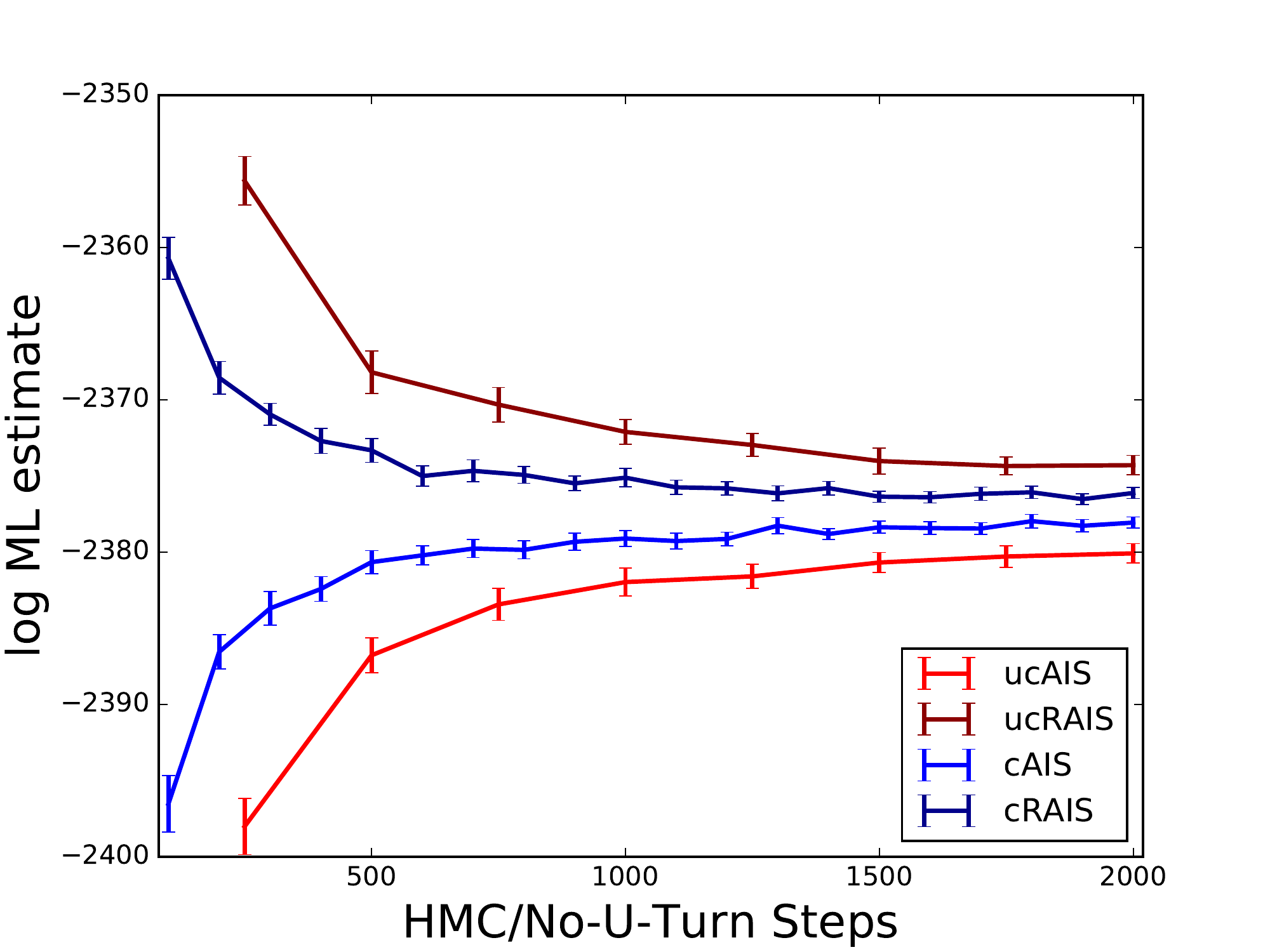}
(b) \includegraphics[width=0.28 \textwidth]{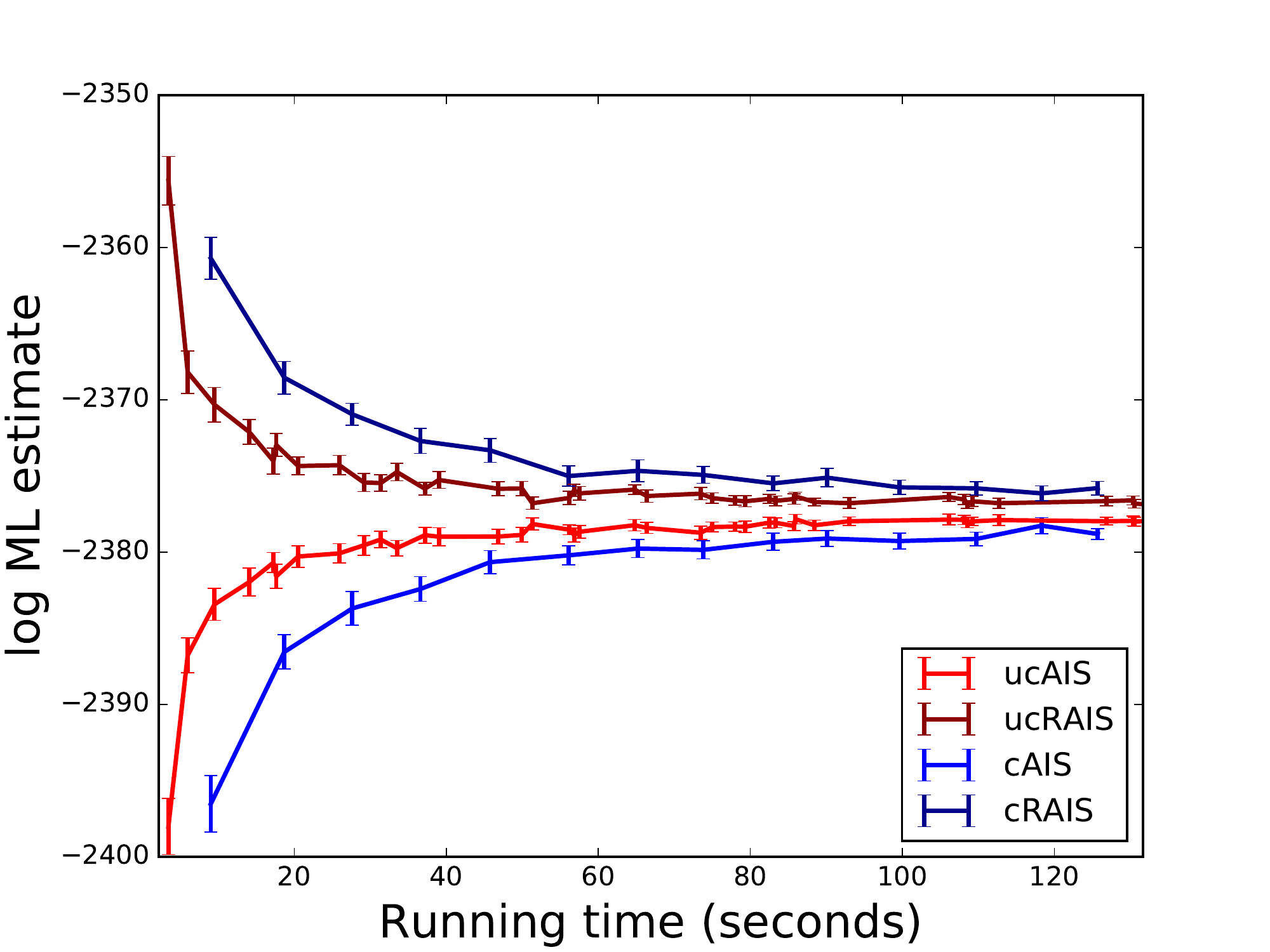} 
(c) \includegraphics[width=0.28 \textwidth]{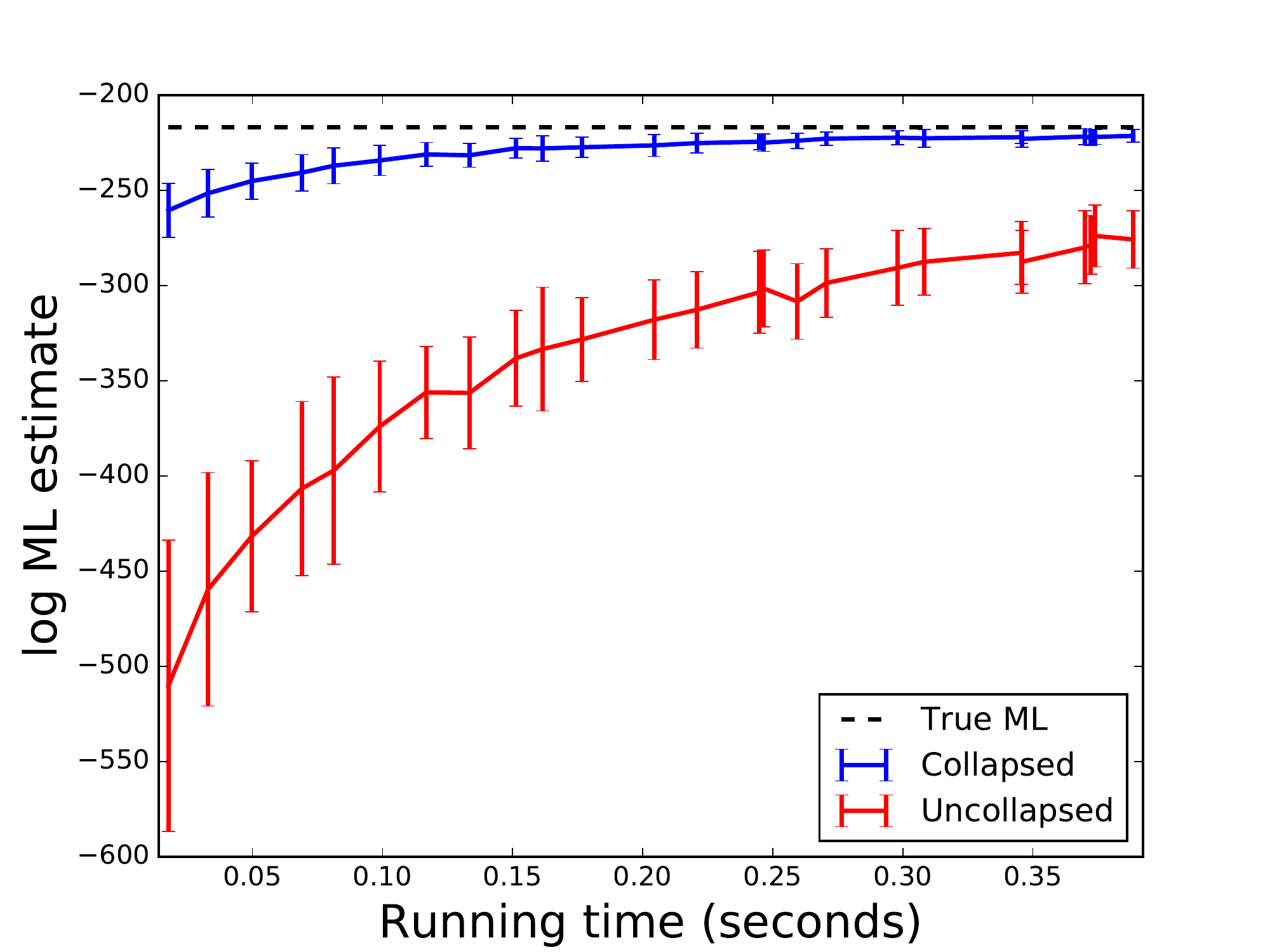}
\end{small}
\end{center}
\caption{AIS and reverse AIS for matrix factorization model in Stan and WebPPL.  {\bf (a-b)} Log-ML estimates for Stan as a function of number of MCMC steps (a) and running time (b). Convergence is faster for the collapsed model in terms of steps, but faster for the uncollapsed model in terms of running time.  {\bf (c)} Log-ML stochastic lower bounds for WebPPL as a function of number of running time. The dashed line is the log-ML upper bound as estimated in Stan. In contrast with Stan, the collapsed version converges much faster than the uncollapsed one. }
\label{mf}
\end{figure}

We consider the matrix factorization model, where we approximate an $N \times D$ matrix $\mathbf{Y}$ as a low rank matrix, the product of matrices $\mathbf{U}$ and $\mathbf{V}$ with dimensions $N\times K$ and $K\times D$ respectively (where $K < \text{min}(N, D)$). We use a spherical Gaussian observation model, and spherical Gaussian priors on $\mathbf{U}$ and $\mathbf{V}$.
\begin{equation*}
u_{ik} \sim \mathcal{N}(0, \sigma_{u}^{2}) \qquad
v_{kj} \sim \mathcal{N}(0, \sigma_{v}^{2}) \qquad
y_{ij} \sim \mathcal{N}(\mathbf{u}_{i}^{T} \mathbf{v}_{j}, \sigma^{2})
\end{equation*}
We can collapse $\mathbf{U}$ and rewrite this model as
\begin{equation*}
v_{kj} \sim \mathcal{N}(0, \sigma_{v}^{2}) \qquad
\mathbf{y}_{i} \sim \mathcal{N}(\mathbf{0}, \sigma_{u}\mathbf{V}^{T}\mathbf{V} + \sigma \mathbf{I})
\end{equation*}
We fix the values of all hyperparameters to 1, and set $N=50$, $K=5$ and $D=25$. We ran \toolname\ on this model in Stan and plotted the BDMC curves (see \cref{mf}).

In general, collapsing variables can result in faster convergence (in terms of number of iterations) at the expense of slower updates. Consistent with this story, inference appears to converge faster on the collapsed model converges in terms of the number of MCMC steps. However, there is greater computational overhead per MCMC step in the collapsed version, since it involves sampling from a multivariate Gaussian distribution. Overall, the tradeoff favors the uncollapsed version in terms of running time. Hence \toolname\ suggests that the user need not collapse the matrix factorization model in Stan.

However, the story is different in the case of WebPPL. \cref{mf} shows AIS curves for collapsed and uncollapsed versions, for a smaller-sized matrix factorization model where $N=10$, $K=5$ and $D=10$. The true log-ML value was obtained by running \toolname\ in Stan where the BDMC bounds had converged to within 1 nat.  \toolname\ on WebPPL shows that the collapsed version enables far faster convergence in terms of running time; therefore, unlike with Stan, it is advisable to collapse the model in WebPPL. Hence \toolname\ can provide insight into the tricky question of which representations of models to choose to achieve faster convergence.

\subsubsection{Debugging}

Mathematically, the forward and reverse AIS chains yield lower and upper bounds on $\log \pmf(\obs)$ with high probability; if this behavior is not observed, that indicates a bug. In our experimentation with WebPPL, we observed a case where the reverse AIS chain yielded estimates significantly lower than those produced by the forward chain, inconsistent with the theoretical guarantee. This led us to find a subtle bug in how WebPPL sampled from a multivariate Gaussian distribution (which had the effect that the exact posterior samples used to initialize the reverse chain were incorrect).\footnote{Issue: \url{https://github.com/probmods/webppl/issues/473}}  These days when many new probabilistic programming languages are emerging and many are in active development, such debugging capabilities provided by \toolname\ can potentially be very useful.

\small
\printbibliography

\appendix

\section{Importance sampling analysis of AIS}
\label{app:ais_derivation}

This is well known and appears even in \citep{AIS}.  We present it here in order to provide some foundation for the analysis of the bias of the log marginal likelihood estimators and the relationship to the Jeffreys divergence.

\citet{AIS} justified AIS by showing that it is a simple importance sampler over an extended state space:
To see this, note that the joint distribution of the sequence of states $\genState_{1}, \dotsc, \genState_{\ndist-1}$ visited by the algorithm has joint density
\[
\pmfProposalForward(\genState_1, \dotsc, \genState_{\ndist-1}) 
&= \pmf_1(\genState_1)\, \trans_2(\genState_2 \given \genState_1) \dotsm \trans_{\ndist-1}(\genState_{\ndist-1} \given \genState_{\ndist - 2}),
\]
and, further note that, by reversibility,
\[
\trans_{\distIdx} ( \genState' \given \genState) 
= \trans_{\distIdx} (\genState \given \genState') 
                \frac { \pmf_{\distIdx} (\genState') }
                        { \pmf_{\distIdx} (\genState) }                
= \trans_{\distIdx} (\genState \given \genState') 
                \frac { \pmfUnnorm_{\distIdx} (\genState') }
                        { \pmfUnnorm_{\distIdx} (\genState) }   .            
\]
Then, applying \cref{eqn:ais-update}
\[
\expect[\weight] &=
\expect\biggl[\, \prod_{\distIdx=2}^{\ndist} 
                \frac{\pmfUnnorm_{\distIdx}(\genState_{\distIdx-1})}
                       {\pmfUnnorm_{\distIdx - 1}(\genState_{\distIdx-1})} \biggr]
\\&= 
\expect
   \biggl[\, 
    \frac{\pmfUnnorm_\ndist(\genState_{\ndist-1})}
           {\pmfUnnorm_1(\genState_1)} 
    \frac{\pmfUnnorm_2(\genState_1)} 
           {\pmfUnnorm_2(\genState_2)} 
         \dotsm 
     \frac{\pmfUnnorm_{\ndist-1}(\genState_{\ndist-2})}
            {\pmfUnnorm_{\ndist-1}(\genState_{\ndist-1})}
    \biggr]
\\&= 
\frac{\pfn_\ndist}{\pfn_1}
\expect
   \biggl[\, 
    \frac{\pmf_\ndist(\genState_{\ndist-1})}
           {\pmf_1(\genState_1)} 
    \frac{\trans_2(\genState_1 \given \genState_2)}
           {\trans_2(\genState_2 \given \genState_1)} 
       \dotsm 
    \frac{\trans_{\ndist-1}(\genState_{\ndist-2} \given \genState_{\ndist-1})}
           {\trans_{\ndist-1}(\genState_{\ndist-1} \given \genState_{\ndist-2})}
   \biggr]
\\&= 
\frac{\pfn_\ndist}{\pfn_1}
\expect
   \biggl[\, 
    \frac{ \pmfProposalBackward(\genState_1, \dotsc, \genState_{\ndist-1}) }
           { \pmfProposalForward(\genState_1, \dotsc, \genState_{\ndist-1}) }
   \biggr]
\\&= 
\frac{\pfn_\ndist}{\pfn_1},
\]
where
\[
\pmfProposalBackward(\genState_1, \dotsc, \genState_{\ndist-1}) 
&= \pmf_\ndist(\genState_{\ndist - 1})\, \trans_{\ndist-1}(\genState_{\ndist-2} \given \genState_{\ndist-1})  
           \dotsm \trans_2(\genState_1 \given \genState_2)
\]
is the density of a hypothetical \emph{reverse} chain,
where $\genState_{\ndist-1}$ is first sampled exactly from the distribution $\pmf_\ndist$, and the transition operators are applied in the reverse order. (In practice, the intractability of $\pmf_\ndist$ generally prevents one from simulating the reverse chain.)

\end{document}

%% file: commenting.tex
\definecolor{WowColor}{rgb}{.75,0,.75}
\definecolor{SubtleColor}{rgb}{0,0,.50}

\newcounter{margincounter}

%% file: macros.tex
\newcommand{\expect}{\mathbb{E}}

\newcommand{\normal}{{\cal N}}

\newcommand{\kldiv}{{\mathrm{D}_{\rm KL}}}
\newcommand{\klBars}{{\,\|\,}}
\newcommand{\given}{{\hspace{.1em}|\hspace{.1em}}}

\newcommand{\subjectiveDivergence}{\mathrm{D}_{\mathrm{SBJ}}}
\newcommand{\symmKLUpperBoundEstimate}{\hat{\mathcal{B}}}
\newcommand{\symmKLUpperBound}{\mathcal{B}}
\newcommand{\symmKLTrue}{\mathcal{J}}
\newcommand{\genUpperBound}{\mathcal{U}}
\newcommand{\genLowerBound}{\mathcal{L}}

\newcommand{\nsamp}{{K}}
\newcommand{\ndist}{{T}}

\newcommand{\sampleIdx}{{k}}
\newcommand{\somethingS}[2]{{#1^{(#2)}}}

\newcommand{\distIdx}{{t}}

\newcommand{\pmf}{{p}}

\newcommand{\pmfUnnorm}{{f}}
\newcommand{\pfn}{\mathcal{Z}}
\newcommand{\pfnRatioFwd}{\mathcal{R}}
\newcommand{\pfnRatioEstimateFwd}{\hat{\mathcal{R}}}
\newcommand{\pfnRatioRev}{\mathcal{R}_{\rm rev}}
\newcommand{\pfnRatioEstimateRev}{\hat{\mathcal{R}}_{\rm rev}}
\newcommand{\pmfProposal}{{q}}

\newcommand{\genState}{\mathbf{x}}
\newcommand{\genStateSpace}{\mathcal{X}}
\newcommand{\genStateS}[1]{{\somethingS{\genState}{#1}}}
\newcommand{\genStateUni}{{x}}

\newcommand{\obs}{\mathbf{y}}

\newcommand{\params}{{\boldsymbol \theta}}

\newcommand{\hyperParams}{\boldsymbol{\eta}}
\newcommand{\hyperParamsReal}{\hyperParams_{\rm real}}

\newcommand{\latent}{\mathbf{z}}

\newcommand{\data}{\mathcal{D}}

\newcommand{\weight}{{w}}
\newcommand{\weightS}[1]{{\somethingS{\weight}{#1}}}

\newcommand{\pmfProposalForward}{{\pmfProposal_{fwd}}}
\newcommand{\pmfProposalBackward}{{\pmfProposal_{rev}}}
\newcommand{\invTemp}{{\beta}}
\newcommand{\trans}{\mathcal{T}}

\newcommand{\auxiliary}{\mathbf{v}}

\newcommand{\statGen}{{h}}

